\documentclass{article}

\usepackage{url}
\usepackage{arxiv}
\usepackage{rotating}

\usepackage[utf8]{inputenc} 
\usepackage[T1]{fontenc}    
\usepackage{hyperref}       
\usepackage{url}            
\usepackage{booktabs}       
\usepackage{amsfonts}       
\usepackage{nicefrac}       
\usepackage{microtype}      
\usepackage{graphicx}
\graphicspath{ {./images/} }
\usepackage{pdflscape}
\usepackage{longtable}
\usepackage{multirow}
\usepackage{multicol}
\usepackage{array}
\title{AI-based Identity Fraud Detection: A Systematic Review}

\author{
 Chuo Jun Zhang \\
  School of Computer Science\\
    University of Technology Sydney \\
  NSW 2007, Australia \\
  \texttt{chuojun.zhang@student.uts.edu.au} \\
   \And
 Asif Q. Gill \\
  School of Computer Science\\
    University of Technology Sydney \\
  NSW 2007, Australia \\
  \texttt{asif.gill@uts.edu.au} \\
  \And
 Bo Liu \\
  School of Computer Science\\
    University of Technology Sydney \\
 NSW 2007, Australia \\
  \texttt{bo.liu@uts.edu.au} \\
   \AND
   Memoona Anwar \\
   School of Computer Science \\
     University of Technology Sydney \\
   NSW 2007, Australia \\
   \texttt{memoona.anwar@uts.edu.au} \\
}

\begin{document}
\maketitle
\begin{abstract}
With the rapid development of digital services, a large volume of personally identifiable information (PII) is stored online and is subject to cyberattacks such as Identity fraud. Most recently, the use of Artificial Intelligence (AI) enabled deep fake technologies has significantly increased the complexity of identity fraud. Fraudsters may use these technologies to create highly sophisticated counterfeit personal identification documents, photos and videos. These advancements in the identity fraud landscape pose challenges for identity fraud detection and society at large. There is a pressing need to review and understand identity fraud detection methods, their limitations and potential solutions. This research aims to address this important need by using the well-known systematic literature review method. This paper reviewed a selected set of 43 papers across 4 major academic literature databases. In particular, the review results highlight the two types of identity fraud prevention and detection methods, in-depth and open challenges. The results were also consolidated into a taxonomy of AI-based identity fraud detection and prevention methods including key insights and trends. Overall, this paper provides a foundational knowledge base to researchers and practitioners for further research and development in this important area of digital identity fraud.  
\end{abstract}


\section{Introduction}
Identity is used to identify an individual or entity [66-68]. Identity fraud is a criminal act involving the use of another person's identity or the creation of a fake identity for fraudulent purposes [69-70]. Fraudsters' main goal or motive could be to obtain money and public resources, evade obligation or commit other crimes [44]. With the rapid development of digital and online services, a large amount of personally identifiable information (PII) is scattered and generally stored in public databases. The monetary benefits of selling personal information on the black market have driven cyber-criminals to covet this personal data. Every year, billions of PII records are breached or stolen due to cyber-attacks and security vulnerabilities [56-64], providing the means for committing identity fraud. Between 2020 and 2024, multiple data breaches have occurred, compromising personal information such as names, dates of birth, social security numbers, addresses, emails, phone numbers, and even bank account passwords [56-64]. In recent years, the development of Generative Artificial Intelligence (GAI) technology has made identity fraud feasible with the possibility of generating fake identities at a scale. Several fraud cases involving deep fake technology have emerged recently. For instance, deepfake allowed criminals to bypass identity verification systems and conduct fraud through live video [45]. The consequences of identity theft and the use of fake identities are far-reaching, influencing both individuals and organisations across a wide range of sectors [70]. The fraud ring or individuals obtain true identity information from others or fabricate fake identities through black market channels, using them for targeted scams, extortion, insurance and loan fraud, and even participating in illegal activities like money laundering (Fig.1) [44].

AI or GAI can be used for both enabling identity frauds via generating fake identities and fraud detection methods. There is a need to review and consolidate knowledge on this topic to facilitate the understanding of AI and identity fraud. Thus, this paper aims to address the following main research question:

\textbf {RQ: What is known about AI-based identity fraud methods?}

This review paper focused on the study of AI-based detection methods covering identity fraud prevention and detection stages. To the best of our knowledge, there is no similar Systematic Literature Review (SLR) study that examines the AI and integrated identity fraud prevention and detection processes. The main contribution and reflections of this paper are as follows:

\begin{enumerate}  
\item A novel taxonomy for identity fraud prevention and detection methods, categorised into two key phases: authentication and continuous authentication. This taxonomy provides a structured approach to understanding the strategies and challenges employed in each phase.
\item Key insights about the increasing importance of biometric authentication, especially facial recognition.
\item Continuous authentication using user behaviour analysis, which is becoming more important for improving fraud detection after the initial authentication. \end{enumerate}

This paper is organised as follows. Firstly, it discusses the research motivation and method. Secondly, it presents the analysis of the identified fraud detection methods based on the review of selected papers. Thirdly, it discusses the open challenges before concluding.

\section{Research background and motivation}
Identity, which serves to identify an individual and entity, is widely used daily to access various activities such as public services, education, medication, job applications, financial transactions, social media, etc. [66]. The threats to identity are known as identity theft and identity fraud, which are collectively referred to as identity-related crimes, including all forms of illicit conduct involving identity, such as acquiring, manipulating, producing, transferring, possessing, or using identity information for unlawful activities [44], [69]. Identity theft and identity fraud, though related, are not identical. Initially, the two concepts were not clearly distinguished until 2011 when The United Nations Commission on Crime Prevention and Criminal Justice provided a clear definition of them [69],[71]. 
\begin{figure*}[ht] 
    \raggedright
    \includegraphics[width=\textwidth]{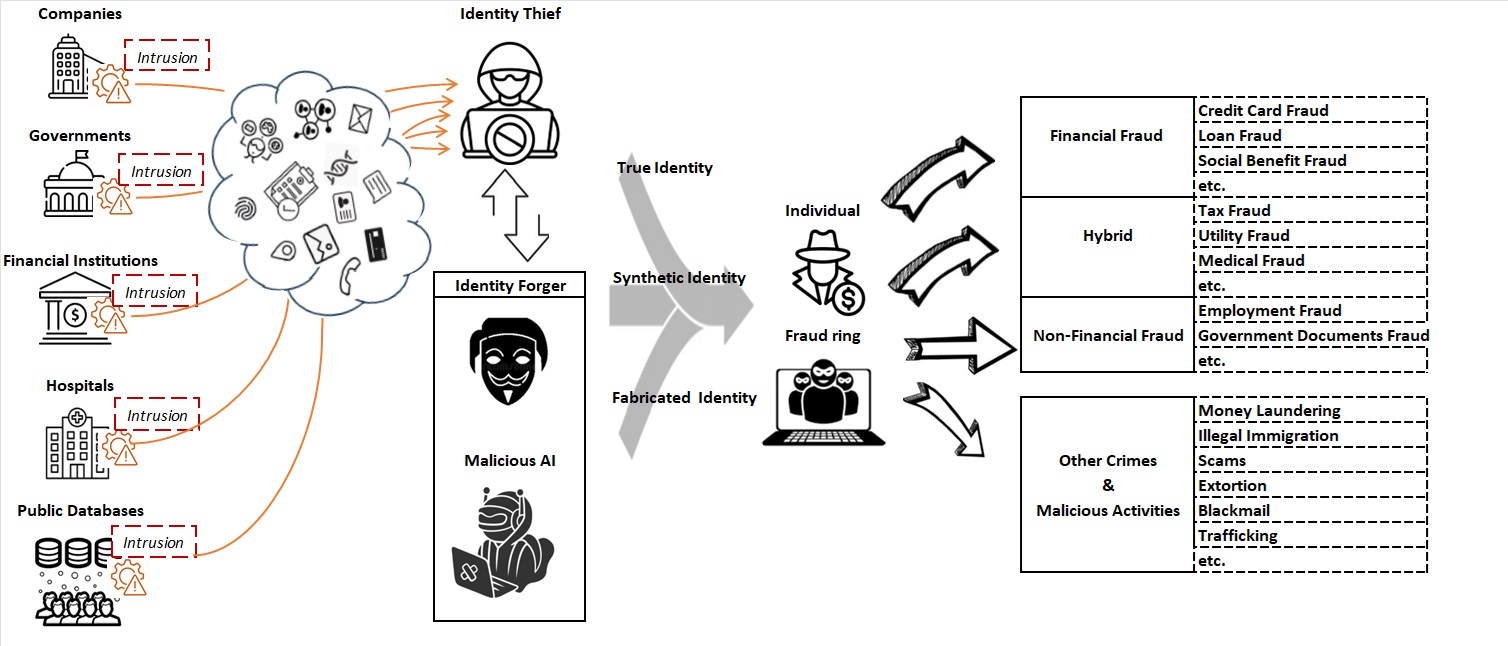}
    \caption{The Identity Fraud Chain: From Data Theft to Fraudulent Use Cases (Image adapted from [44, p59].)} 
    \label{fig:1}
\end{figure*}

Identity theft is often the first step of identity fraud, which primarily involves the unlawful acquisition of identity information, such as PII [78]. As shown in Fig \ref{fig:1}., identity thieves initiate malicious attacks, including malware, phishing, and brute force, to intrude into various institutions such as financial institutions, hospitals, governments, and enterprises to steal identity information [56–64]. These stolen identities are either directly sold or transferred to accomplices, such as identity forgers, who produce synthetic or fabricated identities for resale to criminal individuals or groups [78-79]. With the assistance of AI technologies, particularly GAI, the manipulation and creation of both real and synthetic identities have become significantly easier [80]. Identity fraudsters then use this identity information to create new identities or take over other identities to gain either financial or non-financial benefits at the cost of other accounts [44]. For example, defraud financial institutions to receive payouts or services [81]. This identity information can also be used to facilitate a means to an end as part of other illegal acts such as human trafficking, money laundering, and illegal immigration [82]. Beyond financial losses, the consequences of identity fraud extend, encompassing psychological distress, reputation damage, and potential legal repercussions for victims [72-77]. 

The current defence against identity fraud primarily relies on two processes: authentication and continuous authentication. Authentication is the process of verifying whether someone or something is truly who or what it claims to be. Authentication methods are categorised into three categories based on the type of information being verified: knowledge, possession, and inherence [46]. Knowledge-based authentication involves confirming a user's identity through information they know, such as a one-time password (OTP), answers to security questions, or a personal identification number (PIN) [84]. Possession-based authentication verifies identity through physical items the user owns, such as mobile devices, smart cards, or hardware tokens [83]. Inherence-based authentication confirms a user's identity through their biometric or behavioural characteristics. This type of biometric authentication includes, but is not limited to, facial recognition, voice recognition, and fingerprint recognition [46],[47],[85]. 

However, initial authentication alone may not be enough, as data breaches and advanced attack techniques (e.g., biometric spoofing and deepfakes) still enable criminals to bypass verification systems relatively easily [65],[86-87]. Therefore, defending against identity fraud also requires continuous authentication. Continuous authentication is the process of continuously verifying the user's identity and detecting anomalies by monitoring user behaviour and contextual data in real-time [48]. It reduces vulnerabilities to unauthorised access and enhances the overall security of online interactions and transactions [88].

AI, which originated in 1956 to replicate human thinking and problem-solving [49], has evolved to effectively combat identity fraud in both authentication and continuous authentication. By integrating machine learning and deep learning models with technologies such as computer vision, natural language processing (NLP), and graph technology, AI empowers systems to perform tasks like identification, recognition, and anomaly detection [49]. For example, in biometric authentication, AI leverages deep learning and computer vision technologies to authenticate users by first recognising and then comparing their biometric features[47],[53]. In continuous authentication, AI incorporates graph technology for relationship modelling and NLP for analysing textual data to detect fraud patterns across structured data (e.g., transaction records) and unstructured data (e.g., social media content), which enables real-time and adaptive authentication [48].

Existing literature on identity fraud reveals the following gaps, which this review aims to address. These gaps include (1) many existing reviews primarily address financial fraud detection but not identity fraud detection [93-96];  (2) a lack of exploration into how AI can be applied across both stages of authentication to prevent identity fraud effectively, with existing studies focusing mainly on specific technical aspects of fraud detection[93],[95-96].

This review, motivated by an industrial project to develop an AI fraud engine for identity verification systems, investigates how AI detects identity fraud during authentication processes and identifies gaps in current methods to guide the development of more robust solutions. It introduces a novel framework by adopting authentication processes as a consolidated structure, systematically deconstructing and analysing each phrase of identity fraud detection methods and AI integration. This hierarchical "top-down" approach provides a rigorous and structured exploration of methodologies. Furthermore, this review addresses existing research gaps and provides insights and future directions for researchers and practitioners.

\section{Research Methodology} 
This review conducts a systematic literature review based on the detailed approach described in [51]. It begins with research planning, where we define research questions, keywords, databases, and selection criteria. Then, we conduct the data search and selection based on the inclusion and exclusion criteria and quality assessment criteria. Finally, we extract data from the selected documents according to the defined fields

\subsection{Research Questions}
The primary aim of this SLR is to discover all AI-based methods to detect and prevent identity fraud during authentication processes. The methods are compared to evaluate their effectiveness, and identify open challenges and future development points in combating identity fraud. 

This SLR will address this main research question: 

\textbf{RQ: What is known about AI-based identity fraud methods?}

It has been further divided into the following sub-questions:
\begin{itemize}
\item RQ1: What are the current AI-based identity fraud detection methods?
\item RQ2: What are the key open challenges in addressing identity fraud?
\end{itemize}

\subsection{Search Strategy}
The search strategy includes on selecting the databases,

keywords, search strings, and selection criteria in the search.

Four popular databases were selected because they cover a wide range of high-quality, peer-reviewed technology and computer science articles. The databases are listed as: 
\begin{itemize}
\item ACM Digital Library (https://dl.acm.org/);
\item IEEE Xplore (https://ieeexplore.ieee.org/);
\item ScienceDirect (https://www.sciencedirect.com/);
\item Scopus (https://www.scopus.com/) 
\end{itemize}

Based on the research questions defined in Section II-A, we divide the research questions into three search categories and collect a series of related concepts for each (TABLE \ref{tab:1}). We then formulated the search query as follows:

\begin{table}[ht]
    \centering
   \begin{tabular}{|c|}
     \hline
     \begin{minipage}{16cm} 
     \centering%
     \hspace{1 pt}\\
       \hspace{1 pt}\textit{(“identity” OR “ID”)} \\
       \hspace{1 pt}AND \\
       \hspace{1 pt}\textit{(“fraud” OR “theft” OR “scam”)} \\
       \hspace{1 pt}AND \\
       \hspace{1 pt}\textit{(“detect” OR “prevent” OR “verify” OR “authenticate”)}\\
       \hspace{1 pt}AND \\
       \hspace{1 pt}\textit{(“artificial intelligence “OR “machine learning” OR “deep learning”)}\\ 
     \end{minipage} \\
     \hline       
    \end{tabular}
\end{table}

\begin{table}[ht] 
        \centering
        \caption{Search Categories}
        \label{tab:1}
        \begin{tabular}{|p{5cm}|p{10cm}|}
        \hline
\textbf{Search Categories} & \textbf{Relevant Concepts} \\ \hline
Identity fraud detection & identity, ID, fraud, theft, scam, detect, prevent, verify, authenticate \\ \hline
AI-related technologies & machine learning, deep learning, artificial intelligence \\ \hline
        \end{tabular}
    \end{table}

We searched the titles and abstracts for articles published from 2020 to 2024. Limiting the search data range is driven by the fact that the big jump in the use of AI started from the introduction of GPT-3 by OpenAI in 2020, though AI and its related technologies have been studied for years [52]. 

We defined the following inclusion and exclusion criteria to restrict the selection of articles from the returned results (TABLE \ref{tab:2}). 

\begin{table}[ht]
    \centering
    \caption{Inclusion and Exclusion Criteria}
    \begin{tabular}{|p{1cm}|p{6cm}|p{6cm}|}
        \hline
        \textbf{No.} & 
        \textbf{Inclusion} & \textbf{Exclusion}\\ \hline
        1 & Articles from 2020 to 2024 &
        Book chapters, books, datasets, newspapers, magazines, newsletters \\ \hline
        2 & Peer-reviewed articles &
        Non-computer science articles \\ \hline
        3 & English-language articles &
        Articles on cybercrimes or cyberattacks \\ \hline
        4 & Conference and journal papers &
        Duplicate studies \\ \hline
        5 & Focus on identity fraud &
        \\ \hline
    \end{tabular}
    \label{tab:2}
\end{table}

\begin{figure}[ht]   
    \raggedright
    \includegraphics[scale=0.8]{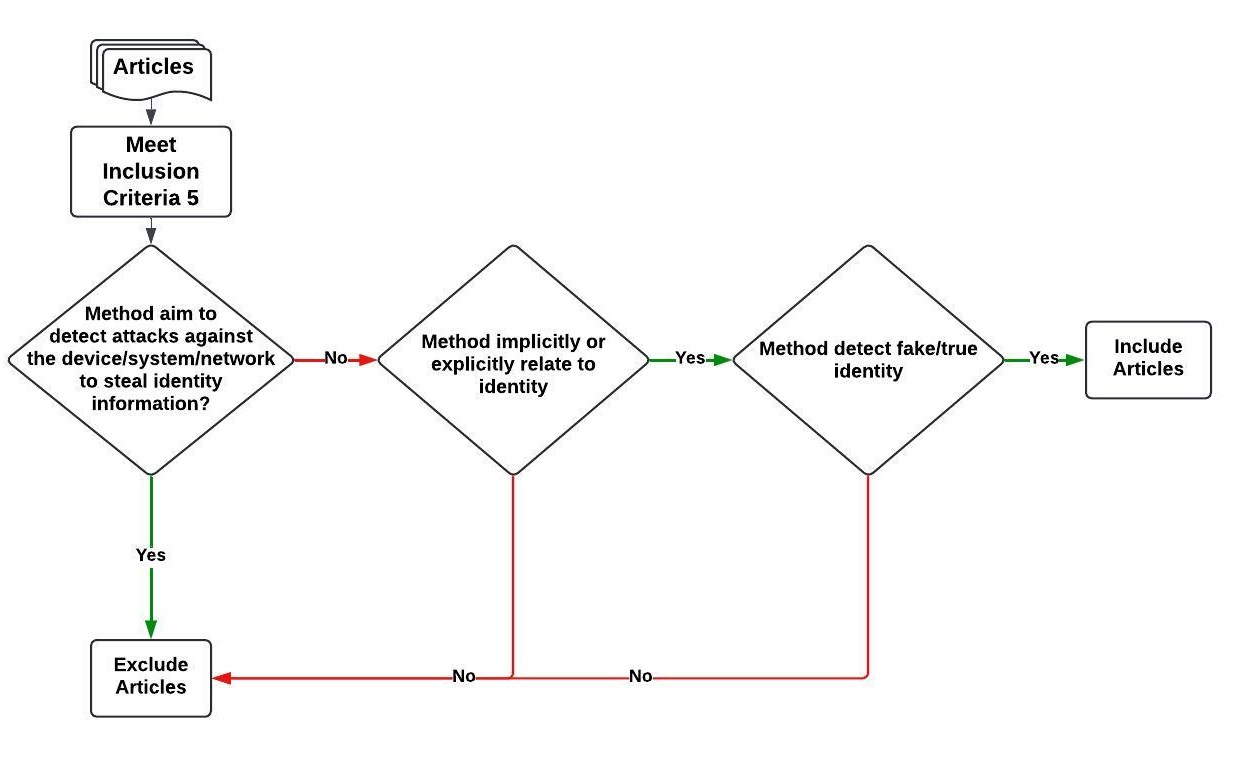}
    \caption{Decision Rules for Inclusion Criteria 5} 
    \label{fig:2}
\end{figure}

This review only includes technical papers on identity fraud detection and excludes articles that address cybercrimes and cyberattacks, as mentioned in Inclusion Criteria 5. The reason for designing Inclusion Criteria 5 this way is that the focus of detecting fraud that involves using false or stolen identities is on verifying the authenticity of the identity being used (i.e., whether it is true or false). However, as the search query includes "identity theft”, articles related to defending against cyberattacks, such as malware detection, are also included in the search results. 
This is not aligned with the research problem of detecting identity fraud. Therefore, to more accurately select articles relevant to this topic, Inclusion Criteria 5 filters the results using the process outlined in Fig. \ref{fig:2}. If an article does not focus on detecting cyberattacks and proposes a method related to "identity" used for verifying true or false identities, it will be included.

The five stages of filtering, as described in TABLE \ref{tab:3}, ensure a thorough selection of relevant articles. First, duplicates are removed, followed by the exclusion of reviews and case studies. Next, papers are filtered based on keywords in their titles and abstracts. The fourth stage applies inclusion and exclusion criteria to refine the selection further. Finally, the remaining full texts are evaluated using six Quality Assessment Criteria (TABLE \ref{tab:4}) to confirm their relevance to the research topic.

\begin{table}[ht]
\centering
\caption{Filtering Rules}
\label{tab:3}
\begin{tabular}{|p{1cm}|p{11.5cm}|}
\hline
\textbf{Filter} & \textbf{Description} \\ \hline
Filter 1 & Duplicate papers \\ \hline
Filter 2 & Remove reviews, surveys, and case studies \\ \hline
Filter 3 & Title includes keywords in the search string \\ \hline
Filter 4 & Title and abstract filtering using exclusion criteria 2 \& 3 and inclusion criteria 5 \\ \hline
Filter 5 & Text filtering \\ \hline
\end{tabular}
\end{table}

\begin{table}[ht]
\centering
\caption{Quality Assessment Criteria}
\label{tab:4}
\begin{tabular}{|c|p{8cm}|}
\hline
\textbf{ID} & \textbf{Quality Assessment} \\ \hline
1 & Does the study clarify its purpose? \\ \hline
2 & Is the study relevant to the research questions? \\ \hline
3 & Are the techniques presented and implemented? \\ \hline
4 & Are the results evaluated adequately? \\ \hline
5 & Does the study provide clear findings and discussion? \\ \hline
6 & Is the future direction clearly stated? \\ \hline
\end{tabular}
\end{table}

This section gathers and extracts data from the primary studies addressing the review questions outlined in Section II-A. The extracted data were organised in a spreadsheet, and the critical information retrieved from these studies was summarised using the Data Extraction Form in TABLE \ref{tab:5}.

\begin{table}[ht]
\centering
\caption{Data Extraction Form}
\label{tab:5}
\begin{tabular}{|p{11cm}|p{3cm}|}
\hline
\textbf{Information Extracted} & \textbf{Research Questions} \\ \hline
The identified problems in identity fraud detection & RQ1 \\ \hline
The techniques used for identity fraud detection & RQ1 \\ \hline
The strengths and weaknesses of the techniques & RQ1 \\ \hline
The AI technologies used in the techniques & RQ1 \\ \hline
The evaluation metrics used to assess the effectiveness of the techniques & RQ1 \\ \hline
The existing challenges, gaps, trends, and future direction of the study & RQ1\&RQ2\\ \hline
\end{tabular}
\end{table}

\subsection{Search Results}
\begin{figure}[ht]
    \raggedright
    \includegraphics[scale=0.8]{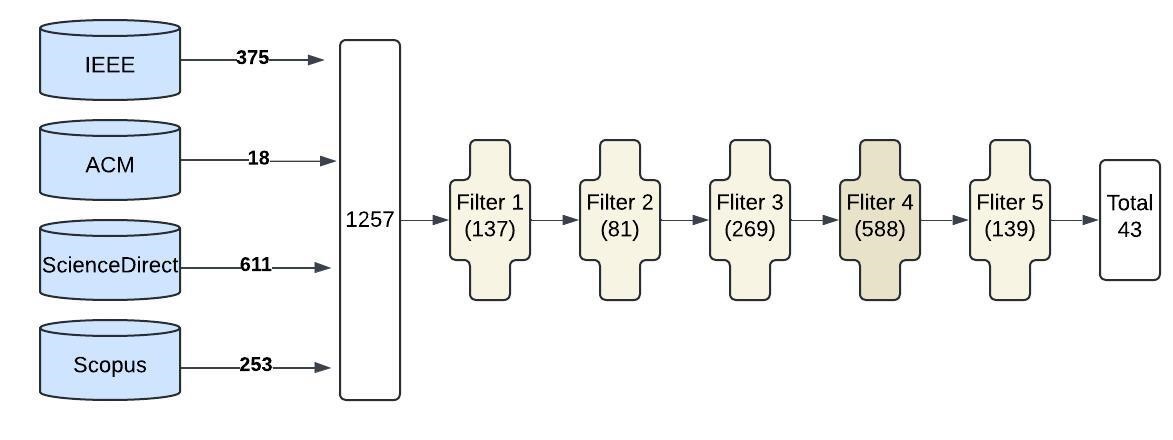}
    \caption{Search Results } 
    \label{fig:3}
\end{figure}

A total of 1257 papers were extracted from all databases, of which 137 were duplicated. The review excluded paper types of survey, literature review, systematic literature review, and case report, in which 81 papers were filtered out by Filter.2. Filter.3 then picked out 269 papers in which keywords were not included in their titles and abstracts. At the same time, Filter 4 further examined the titles and abstracts based on exclusion criteria three and inclusion criteria five, i.e. any papers not related to identity fraud and related to prevention against cyberattacks were excluded. 588 papers were exempted from Filter.4. The quality assessment criteria for the remaining are in TABLE \ref{tab:4}. If the paper meets the requirements, it will score 1; otherwise, it will score 0. Papers in a score range from 3-6 were selected. Finally, 43 papers remained for this review. Fig. \ref{fig:3} presents the selection outcomes, while \ref{appendix:A} provides the scoring details for the chosen papers.

\section{Results}
This section presents the results of the SLR, addressing RQ1 and RQ2. The findings are organised into five main areas:  (A) taxonomy of AI-based identity fraud (IDF) detection methods (Fig. \ref{fig:4}), (B) framework and frequency analysis for open challenges, (C) observed trends, and (D-E)an in-depth analysis of individual methods including the  identification of open challenges across technical and non-technical dimensions.

\subsection{Taxonomy}
\begin{figure*}[ht]  
    \raggedright    \includegraphics[width=\textwidth]{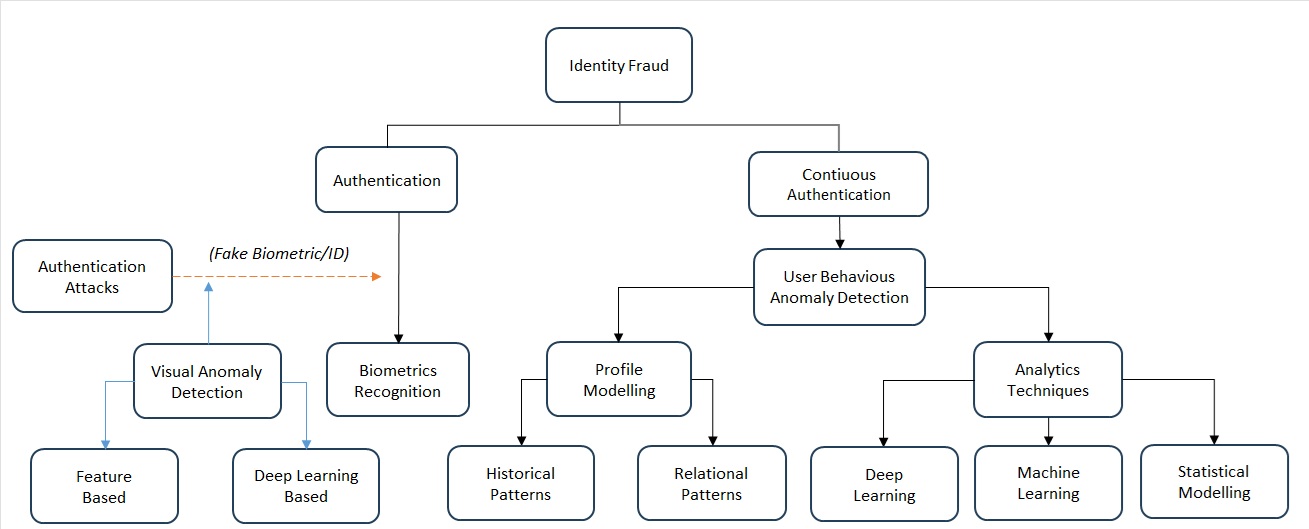}
    \caption{Taxonomy of AI-based Identity Fraud Detection and Prevention Methods} 
    \label{fig:4}
\end{figure*}

The review of the literature resulted in the identification of three main types of AI-based identity fraud (IDF) detection methods: biometric recognition, visual anomaly detection, and user behaviour anomaly detection. These methods are organised into a taxonomy (Fig \ref{fig:4}) and broadly classified into two authentication processes: authentication and continuous authentication.The categorisation was based on the operational purposes, process and core principles of the methods. 

The distinction between authentication and continuous authentication methods lies in their approach to identity verification. Authentication methods typically verify identity at the initial login, while continuous authentication methods involve ongoing monitoring of user behaviour, device characteristics and environmental factors throughout a session [68], [91-92].  
Authentication methods were further classified based on their operational purposes. For example, methods designed to defend against attacks during authentication, such as fake biometrics or identity forgery, were categorised as authentication defence methods. Conversely, methods that solely address authentication system functionality were classified under authentication. Following this second-round classification, a detailed examination of each method’s operational principles, functional components, and implementation techniques was conducted to ensure clear and systematic categorisation.

This review revealed that biometric recognition is the sole method applied for authentication. Methods targeting authentication attacks, such as fake biometrics or identity forgeries, rely on visual anomaly detection, which is further divided into feature-based and deep learning-based approaches. For continuous authentication, all methods fall under user behaviour anomaly detection, where User and Entity Behaviour Analytics (UEBA) serves as the foundational framework. This framework was further examined through two key technical components: profiling models and analytical techniques [91]. The review articles were categorised accordingly to analyse the methods used for constructing behavioural profiles and the analytical techniques applied for anomaly detection, offering a comprehensive understanding of the underlying technologies.

\begin{figure*}[ht]  
    \raggedright
    \includegraphics[width=\textwidth]{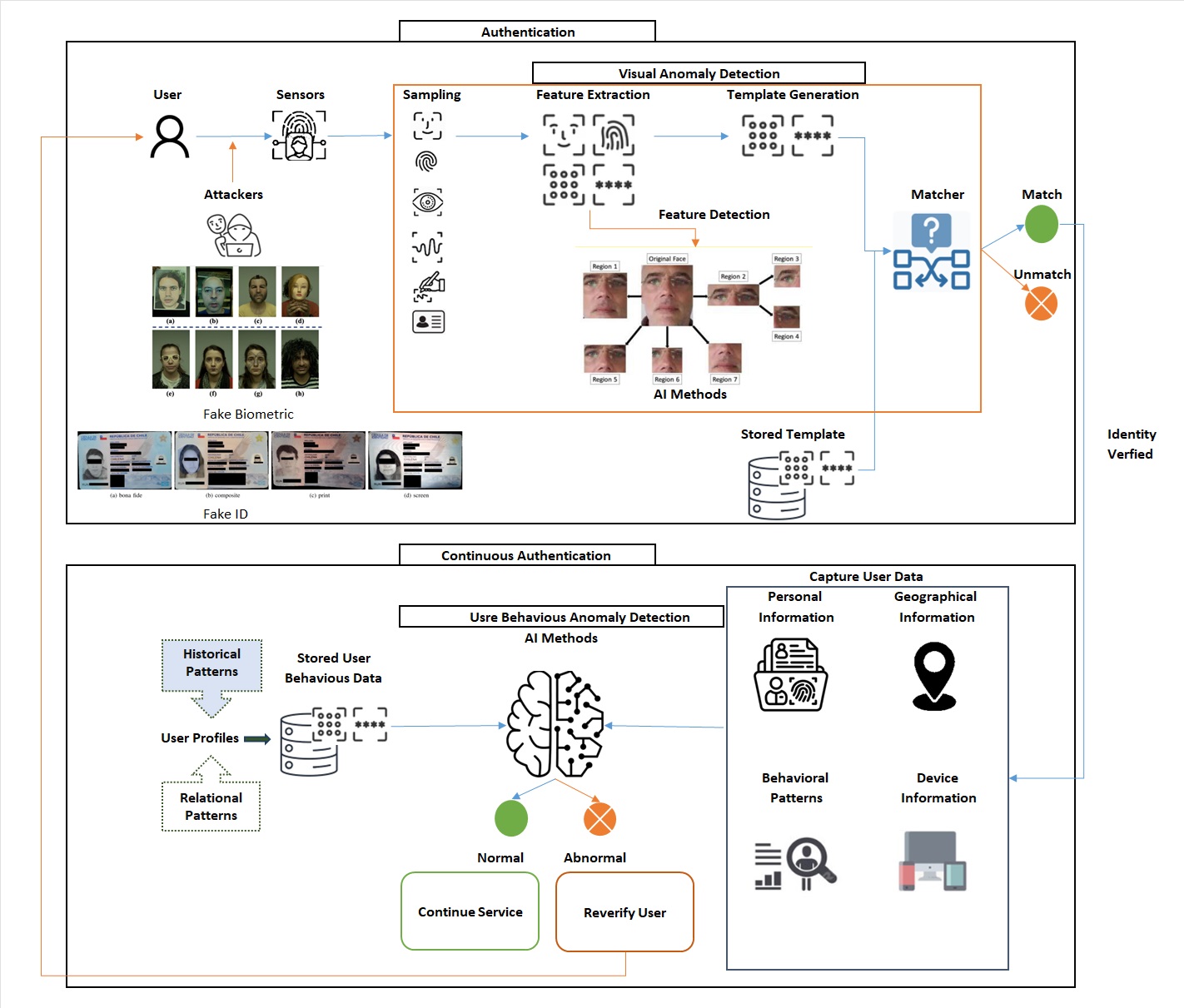}
    \caption{Operational process of authentication and continuous authentication with IDF detection and prevention methods (Source of Image Data: [11], [18] and [65]) 
} 
\label{fig:5}
\footnotesize
\textit{\textbf{Note:}  
\\\textbullet The IDF detection and prevention process begins with identity authentication, where users' biometric data or official ID documents are captured. Key features are extracted to create a user-specific template, which is stored and matched against records to verify identity.\\  \textbullet  Visual anomaly detection with advanced feature extraction algorithms analyses provided samples in detail to safeguard the process. \\ \textbullet  Continuous authentication monitors users' behavioural patterns, relying on baseline profiles to identify anomalies. If anomalies are detected, the system suspends access and prompts for additional identity verification. }
\end{figure*}

Fig. \ref{fig:5} further illustrates how these methods are implemented within two key authentication processes: authentication and continuous authentication. It maps the IDF detection and prevention methods from the taxonomy to their respective roles in these processes, providing a detailed view of their functional implementation.

\subsection{Framework and Frequency Analysis for Open Challenges}

The open challenges identified in this study were examined using the Adaptive Enterprise Architecture (EA) framework [89], which addresses key aspects critical to industrial projects and enterprise decision-making. This structured approach facilitates the categorisation and analysis of challenges in identity fraud detection methods. This review selected three key dimensions from the framework: Technology, Human, and Environment.

The Technology dimension was evaluated based on quality characteristics such as accuracy, robustness, resiliency, and effectiveness [90], while the Human and Environment dimensions were derived and coded to capture non-technical challenges. To quantify and highlight prominent issues across methods, frequency analysis was applied, aligning the findings with key themes identified in the selected literature. This approach was used to examine both individual open challenges for each method and broader challenges across the entire review.

TABLE \ref{tab:13} and TABLE \ref{tab:14} present evaluation criteria and frequency analysis results for the open challenges in this study. The findings highlight four key challenges in AI-based IDF detection methods:  data quality and diversity, data privacy and security, dynamic and evolving fraud patterns, and optimising training time and method efficiency. These challenges will be examined in detail in the discussion section, guiding the exploration of future research directions and advancements.

\begin{table}
    \centering
        \caption{Technology's Challenges and Analysis}
    \begin{tabular}{c}

\includegraphics[width=\textwidth]{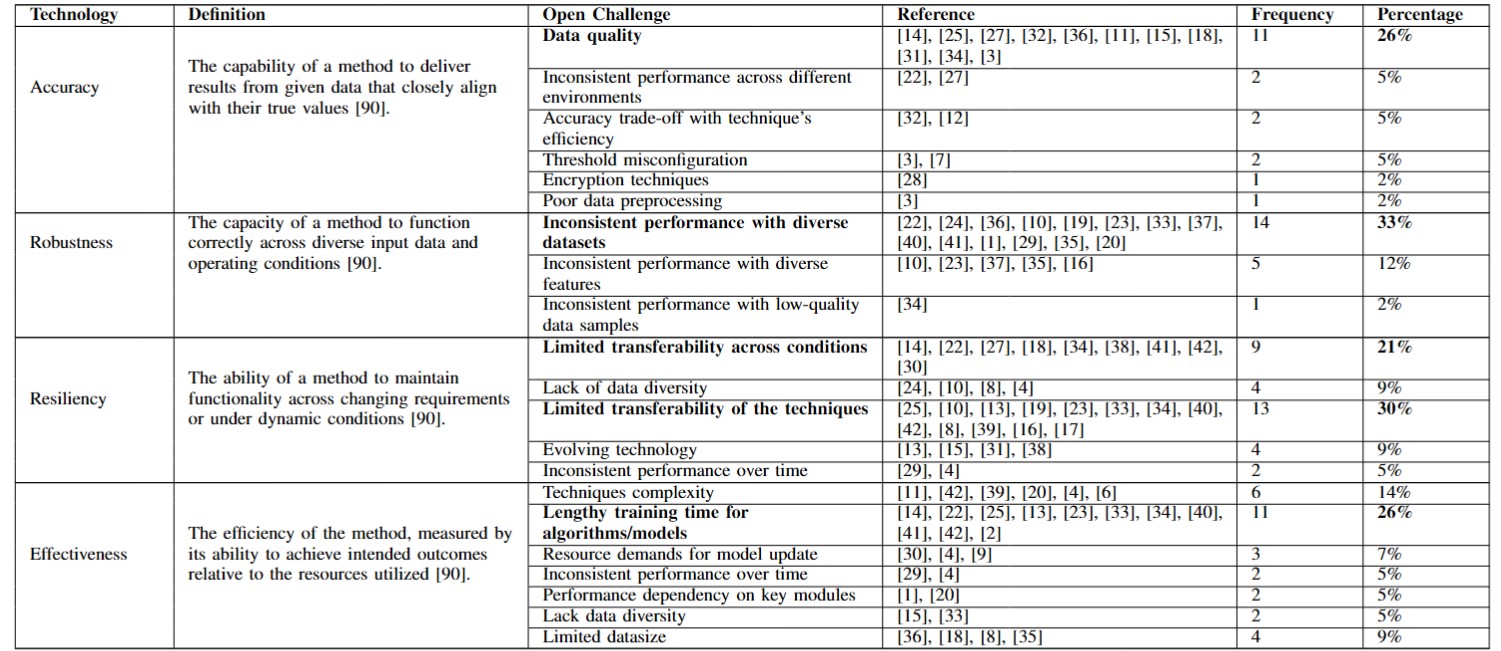} \\

    \end{tabular}

    \label{tab:13}
\end{table}

\begin{table}
    \centering
        \caption{Human's and Environment's Challenges and Analysis}
    \begin{tabular}{c}

\includegraphics[width=\textwidth]{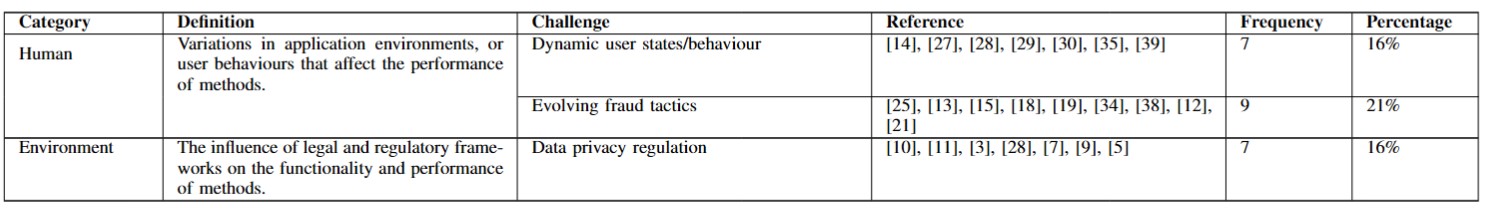} \\

    \end{tabular}

    \label{tab:14}
\end{table}

\subsection{Trends in Method Distribution}
The distribution of AI-based IDF detection methods reveals important trends in the focus and application of these techniques. 
As shown in TABLE \ref{tab:6}, 55.81\% of the methods are predominantly applied to authentication, whereas 44.91 \% are designed for continuous authentication, reflecting a growing emphasis on adaptive and ongoing verification mechanisms.

\begin{figure}
    \centering
    \includegraphics[width=0.7\linewidth]{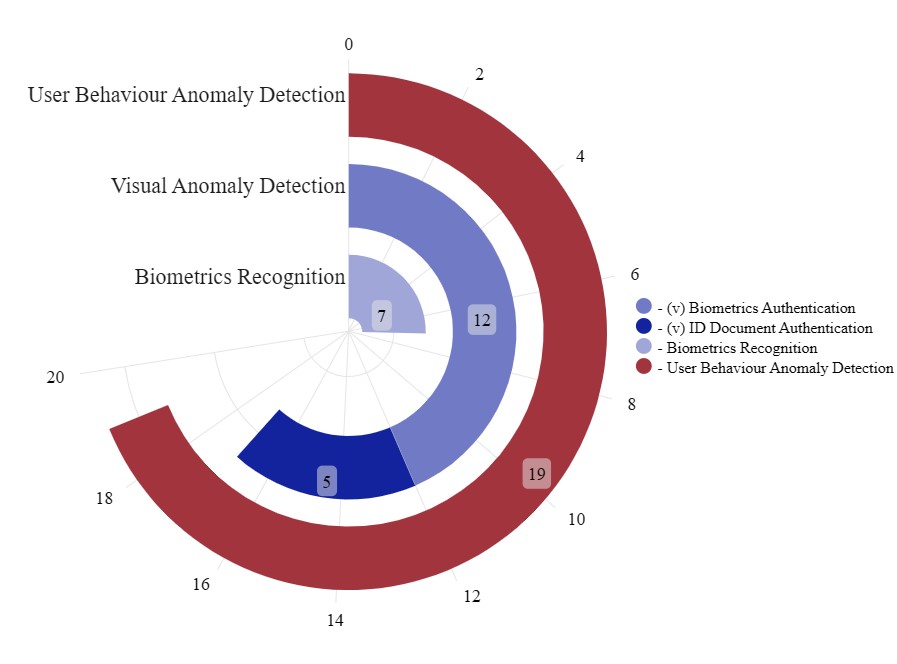}
    \caption{Articles distribution for IDF prevention and detection from 2020 to 2024}
    \label{fig:6}
\end{figure}

As shown in Fig \ref{fig:6} biometric authentication is the most commonly used method in authentication, accounting for 79\% (19/24) of authentication-related articles. Among these, 36.8\% (7/19) focus on performance improvements in biometric recognition systems, while 63.2\% (12/19) address their vulnerabilities. Notably, facial recognition emerges as the most discussed biometric method, appearing in 12 articles within this category.

Visual anomaly detection is identified as the primary defence mechanism in authentication processes, representing 39.35\% of the total articles reviewed. In contrast, continuous authentication methods exclusively rely on user behaviour anomaly detection, which accounts for 44.19\% of the articles.

Deep learning techniques dominate both visual anomaly detection and user behaviour anomaly detection, with 76\% of visual anomaly detection and 53\% of user behaviour anomaly detection articles employing this approach. These findings indicate a growing research emphasis on deep learning as a means to address the evolving challenges of identity fraud.

This distribution provides a foundation for analysing the operational principles, related techniques, and specific challenges
of each method, as elaborated in the subsequent sections.

\begin{table}
    \centering
        \caption{Overview Of Authentication and Continuous Authentication’s IDF Methods}
    \begin{tabular}{c}
\includegraphics[width=0.97\textwidth]{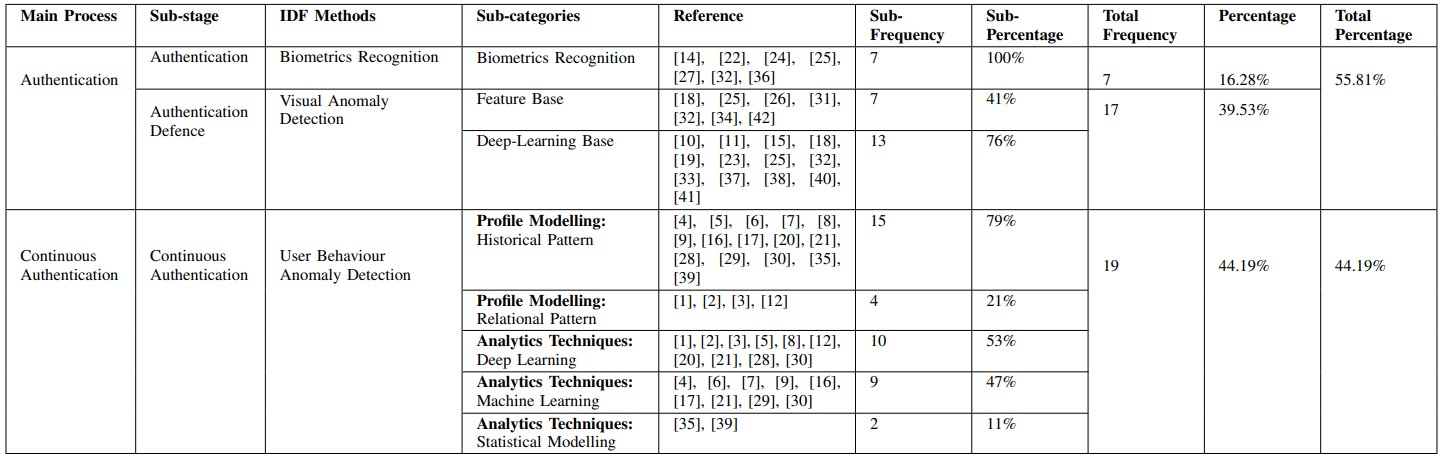} \\

    \end{tabular}

    \label{tab:6}
    \raggedright
    \scriptsize
    \textit{\textbf{Note:} 
    {\\\textbullet The Reference column lists all the papers that include methods within the Sub-categories, including duplicates. \\ \textbullet Sub-Frequency represents the number of papers for each Sub-category. \\ \textbullet Sub-Percentage is calculated as the ratio of Sub-Frequency to Total Frequency, showing the relative contribution of each Sub-category. \\ \textbullet Total Frequency represents the total number of papers across the two Sub-stages. \\ \textbullet Percentage is calculated as Total Frequency divided by the overall frequency of the Main Process (e.g., for Authentication, 7/(7+17)). \\ \textbullet Total Percentage indicates the proportion of the Main Process's total frequency to the total number of selected articles (e.g., for Authentication, 24/43).}}
\end{table}

\begin{figure}
    \centering
    \includegraphics[width=0.7\linewidth]{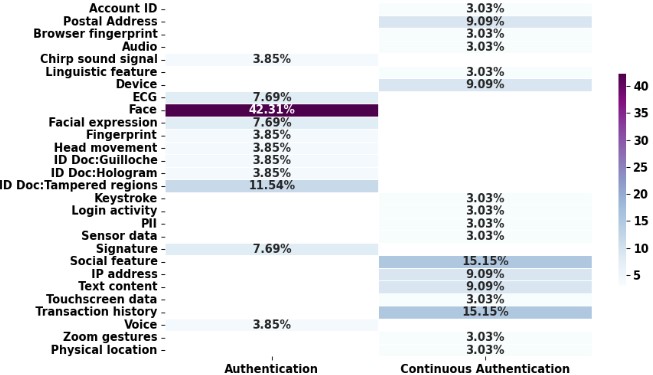}
    \caption{Distribution of Data and Feature Elements in Authentication and Continuous Authentication }
    \label{fig:7}
    \raggedright
    \footnotesize
    \textit{Note: Because some articles used multiple data types, the results reflect an aggregated analysis of the overall data.}
\end{figure}

\subsection{AI-based IDF Prevention and Detection Method: Authentication}

This section focuses on two AI-based authentication methods: biometric recognition and visual anomaly detection. Each method is analysed in detail based on its operational principles, including core components and related implementation techniques. Using the categorisation framework from Section B, the open challenges of current techniques are systematically evaluated and elaborated in the subsequent sections.

\subsubsection{AI-based authentication method: biometrics recognition}
This section analyses the biometric recognition system in terms of its operational components and examines the open challenges identified through frequency analysis.

\paragraph{Operational Principle}

Biometrics refers to the process of identifying individuals by capturing and analysing their unique physical or behavioural characteristics using electronic devices, such as face, fingerprints, signature, voice, Electrocardiogram
(ECG) and chirp Sound [14], [25],[27], [32], [36], [42] as shown in Fig \ref{fig:7}. Biometrics, characterised by their inherent uniqueness and reliability, offer a more effective method for verifying the authenticity of an individual’s identity [27],[53]. As shown in Fig. \ref{fig:5}, a biometric recognition system is composed of the following modules: sensor, feature extraction, matching, and decision-making [53]. The process flow for authentication is depicted by the directional arrows, highlighting the sequence from input data acquisition to final decision-making.
\begin{itemize}
    \item \textit{The Sensor Module} is responsible for capturing a user’s biometric information, like cameras on smartphones, fingerprint scanners, and microphones [27], [41]. 
  \item \textit{The Feature Extraction Module} is used to extract key biometric measures, making the system more effective at identity recognition [42]. For example, facial features such as the eyes, nose, and mouth are crucial for enabling machines to identify individuals. 
are applied to strengthen these features and improve the accuracy of facial recognition systems.
 \item \textit{The Matching Module} is used to compare the newly captured biometric template with the one stored in the database [32]. There are two approaches: one-to-one and one-to-many. In a one-to-one matching system, the biometric template of the verified user is already stored in the system. A common example is a smartphone's identity verification system [22],[24],[25]. In a one-to-many matching, there is no pre-stored biometric template, which means the individual's identity is unknown [32]. The system must search a large database to identify the person. An example of this is the watch list screening conducted by financial institutions before onboarding customers [32].
  \item \textit{The Decision Module} is the system that makes the final decision after the comparison. If the matching result shows a high similarity, the authentication is approved; if the similarity is low, the authentication is denied.
\end{itemize}

\paragraph{Open Challenges}
TABLE \ref{tab:7} summarises the key challenges identified in biometric recognition systems based on the findings of this review. These challenges highlight critical limitations in the current state of the technology, including data quality, system resilience, implementation effectiveness and security vulnerabilities, which are detailed below. 
\begin{itemize}
    \item Data quality is the most frequently cited challenge affecting model accuracy, due to variability in data sampling conditions and sample collection devices (See TABLE \ref{tab:7}). Factors such as lighting and angle [22] are common sources of variability in biometric recognition. As noted in [41], different fingerprint sampling devices produce varying biometric templates, which also increases the difficulty of detection. 
    
    In facial recognition systems, dataset diversity is another critical factor influencing performance. For example, changes in a person’s age or facial expressions in facial recognition systems can impact verification accuracy [32]. If the dataset lacks diverse samples for each user, the verification results may be significantly affected.
    
    \item  High-quality data relies on advanced sampling devices, particularly when employing more specialised biometric features like Chirp Sound and ECG [14], [27], [36]. These features have been validated for their role in biometric recognition [36]. However, implementing these methods requires specialised equipment for capturing high-quality signals. Similarly, noise during voice data collection remains a significant challenge, as highlighted by [27], [43]. 
    \item A significant challenge to achieving the effectiveness of biometric recognition systems lies in their inherent complexity. Long training times [14], [22], [25], intricate biometric data structures such as ECG signal sequences, and extended data collection for specialised biometrics [27] all contribute to delays in achieving practical and timely implementation.
   \item Common biometric systems (e.g., facial recognition, fingerprints, voice recognition) are vulnerable to potential security breaches from evolving fraud technologies, such as morphing attacks and presentation attacks [18], [19], [22], [26], and Deepfake attacks  [13], [15], [23], [37], [40]. Criminals can bypass these systems using simple forgeries, like 2D or 3D printed masks or by displaying 
   images on a phone screen [18], [19]. In high-security areas such as border control, criminals may use morphing techniques by submitting a fused-feature passport photo during visa applications [22]. This changes the stored template in the database, complicating the detection of fraudulent activities. 

\end{itemize}

\begin{table}
    \centering
        \caption{Open Challenges in Biometric Recognition}
    \begin{tabular}{c}

\includegraphics[width=0.97\textwidth]{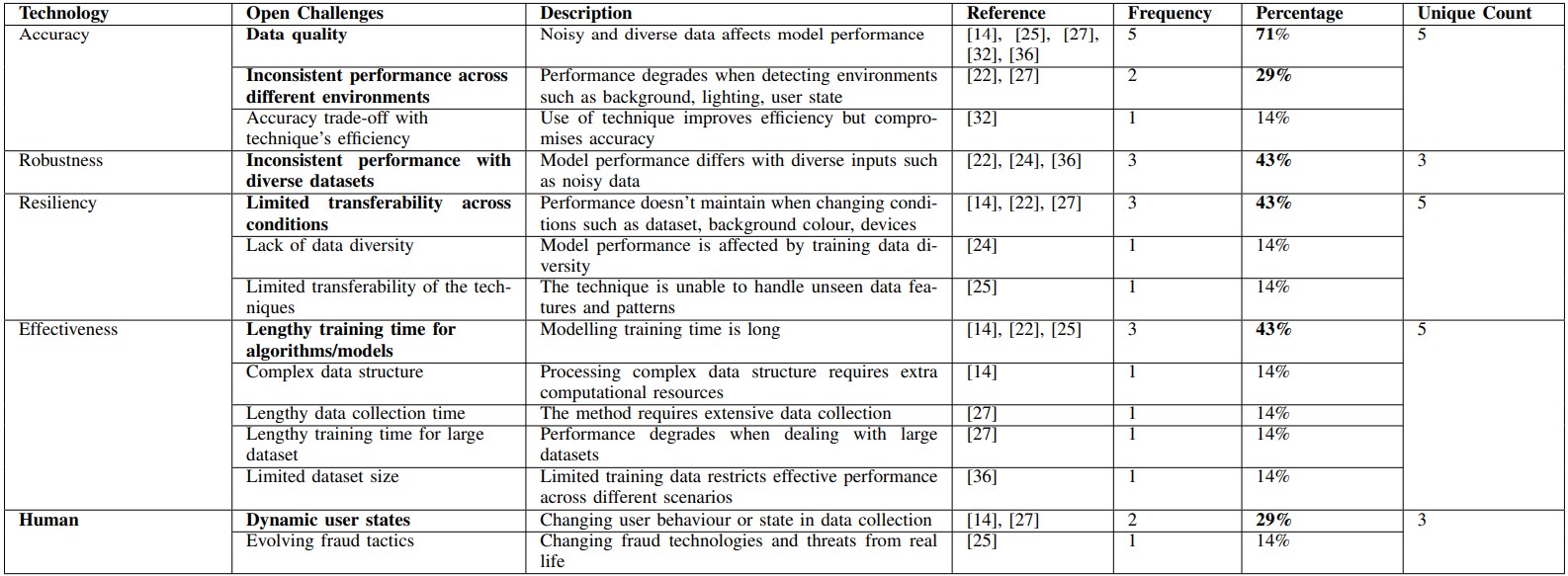} \\

    \end{tabular}
\raggedright
\footnotesize
\textit{\textbf{Note:}{Unique Count refers to the distinct number of references for each criterion.}}
    \label{tab:7}
\end{table}

\subsubsection{AI-based authentication defence method: visual anomaly detection}

This section analyses the visual anomaly detection system in terms of its key implementation techniques. Additionally, it summarises the detection techniques, targeted features, addressed attack types, and the algorithms or models in table and examines the open challenges identified through frequency analysis to provide insights into the limitations of current methods.
\paragraph{Operational Principle}
Visual anomaly detection is a method designed to detect abnormal or unexpected patterns in images and videos that are different from the expected form as defined [54]. This technique is implemented within recognition systems to enhance the matching accuracy with more detailed features, as highlighted in the designated region of the authentication process (Fig \ref{fig:5}). Among the articles discussing vulnerabilities in authentication systems, their main focus is to counter biometric spoofing and deepfake and ID document forgery attacks. TABLE \ref{tab:8} summarised articles for visual anomaly detection along with the types of attacks they counter. Aside from [13], [18] and [19], which use adversarial learning to counter these attacks, other studies [26], [32], [37], [38], [40] rely on techniques in feature detection or deep learning models to achieve successful defence.

Based on the collected articles, visual anomaly detection can be implemented using feature detection and deep learning.
\begin{itemize}
    \item \textit{Feature detection:} Feature detection is a concept in computer vision and image processing. This process uses computers to  extracts information from images and then determines if certain points within the image are part of an object [55]. For instance, feature detection extracts interest points from an image (e.g., points, lines, edges) along with their neighbour area to form regions of interest. By comparing these regions, feature matching can detect global or local anomalies between normal and abnormal images.
    
    Feature detection occurs at the stage of feature extraction to identify abnormal features in images or videos. 
When countering forgery attacks, feature detection enables models to focus on critical areas of change which can effectively detect anomalies, such as eyes and mouths in paper-cut attacks [19], facial expression regions in deepfake videos or subtle statistical differences in morph attacks. Methods like the EBHOG descriptor in [18], [42] and the Multi LBP Model in [34] use feature weighting and multi-feature combination to achieve this.
    
    Conventional feature extraction methods include HOG, Scale-Invariant Feature Transform (SIFT)[25], Local Binary Patterns (LBPs) [34], ORB (Oriented FAST and Rotated BRIEF) [25], and Gabor Filters [32]. In addition to these texture features, statistical features like Multidimensional Co-occurrence Matrices (MCMs) [26] are used to capture spatial relationships of pixel values in an image.

\item \textit{Deep learning:} Deep learning based approach is the more advanced version of feature detection. The effectiveness of deep learning techniques in image classification has motivated researchers to use deep neural networks like Convolutional Neural Network (CNN), Residual Network (ResNet), U-shaped Convolutional Neural Network (U-Net), and Visual Geometry Group Network (VGG)  [23], [25],  [26], [32], [33] for extracting more advanced features from images automatically. CNN [18], [25], [32] and its related architectures, such as Inception and its variant XceptionNet [15], [37], [40], are among the most widely used deep neural networks in visual anomaly detection. For example, methods [15] and [18] use encoder-decoder architectures to extract and segment facial features in detail and reconstruct anomalous samples, allowing the decoder to determine authenticity. Neural networks specifically designed to extract facial features, such as  Viola-Jones Object Detection and Multi-task Cascaded Convolutional Networks [23], [37], are also applied during the training process, particularly targeting complex attacks like face-swap, face replacement, and face reenactment. Additionally, it incorporates feature detection methods, using descriptors that extract key features [18] or the statistical features of the images to enhance the neural network's detection of anomalous features [40].
\end{itemize}
\paragraph{Open Challenges} TABLE \ref{tab:9} provides an overview of the open challenges of the Visual anomaly detection method identified in this review. This method faces several critical challenges, including data quality issues, limited robustness and resilience, operational inefficiencies, data privacy constraints, and increasingly sophisticated fraud tactics driven by technological advancements, as shown below:

\begin{itemize}

    \item Detection accuracy is influenced by the quality of the data such as image resolution [15], and the quality of data processing [11], [31], [34]. For instance, [11] applies generative adversarial networks (GAN), StyleGAN2, and CycleGAN to generate synthetic ID cards. Though the method performs well on trained datasets, the synthetic data lacks real-world fidelity, compromising the model's performance in real-world scenarios. [34] then highlights the impact of document misalignment during video processing, which leads to detection errors.
   
   \item Lack of data diversity adversely affects robustness and transferability, hindering the method's performance in real-world scenarios [10], [19], [23], [33], [40], [41], [42]. As mentioned in [10], [40], [41], [42], the model can only distinguish seen patterns in the trained images and may fail to adapt to diverse scenarios, such as real-time sample capturing with complex lighting, angles, backgrounds and etc.
     \item The use of deep learning methods results in a slower process during both training and testing phases, particularly when handling large datasets [13], [15], [18], [19], [23]. 
  \item  Data privacy regulations have restricted access to private data, affecting the implementation and effectiveness of methods. Both [10] and [11] highlight the prolonged data collection process for real ID document samples due to regulatory constraints. 
  \item Fast-evolving fraud tactics, such as the use of advanced technologies like deepfake, also challenge the ability of these methods to effectively handle new scenarios [13], [15], [31], [38]. For instance, while [11] demonstrates promising performance on previously unseen deepfake models, the high transferability of evolving adversarial examples across diverse model architectures and real-world conditions poses significant challenges.

\end{itemize}

\begin{table}
    \centering
        \caption{Visual Anomaly Detection Method's Article Summary}
    \begin{tabular}{c}
\includegraphics[width=0.97\textwidth]{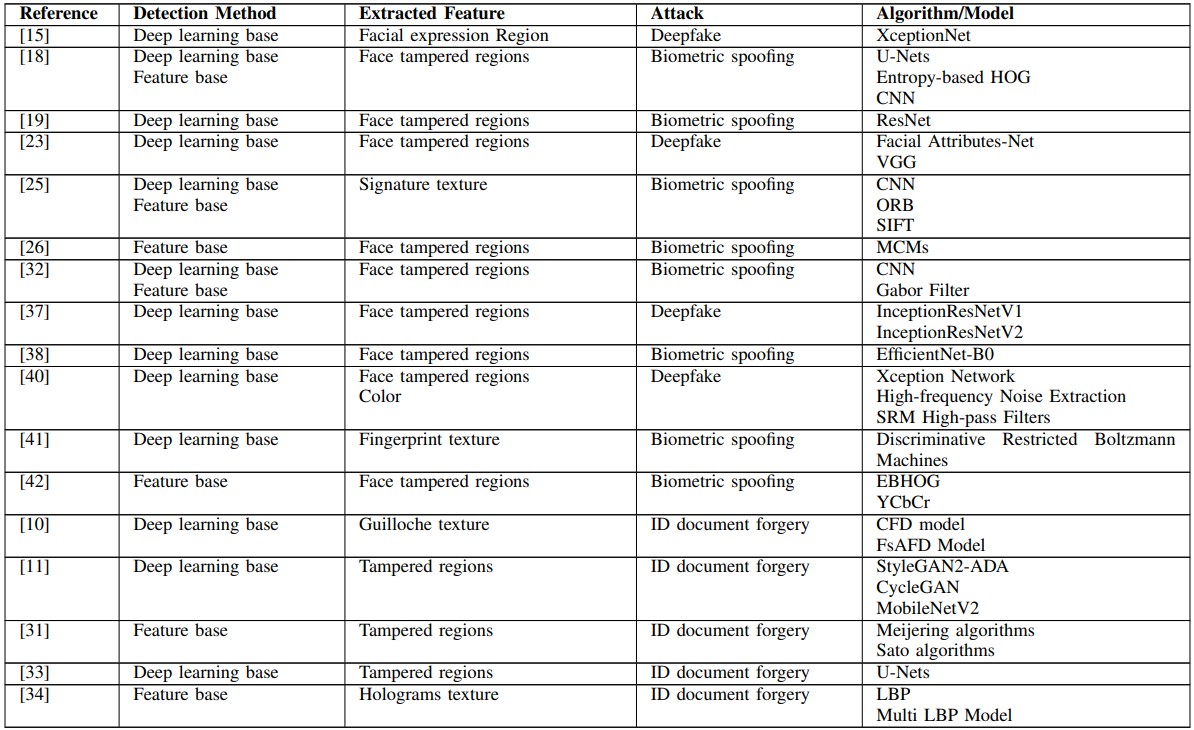} \\

    \end{tabular}

    \label{tab:8}
\end{table}

\begin{table}
    \centering
        \caption{Open Challenge in Visual Anomaly Detection}
    \begin{tabular}{c}
\includegraphics[width=0.97\textwidth]{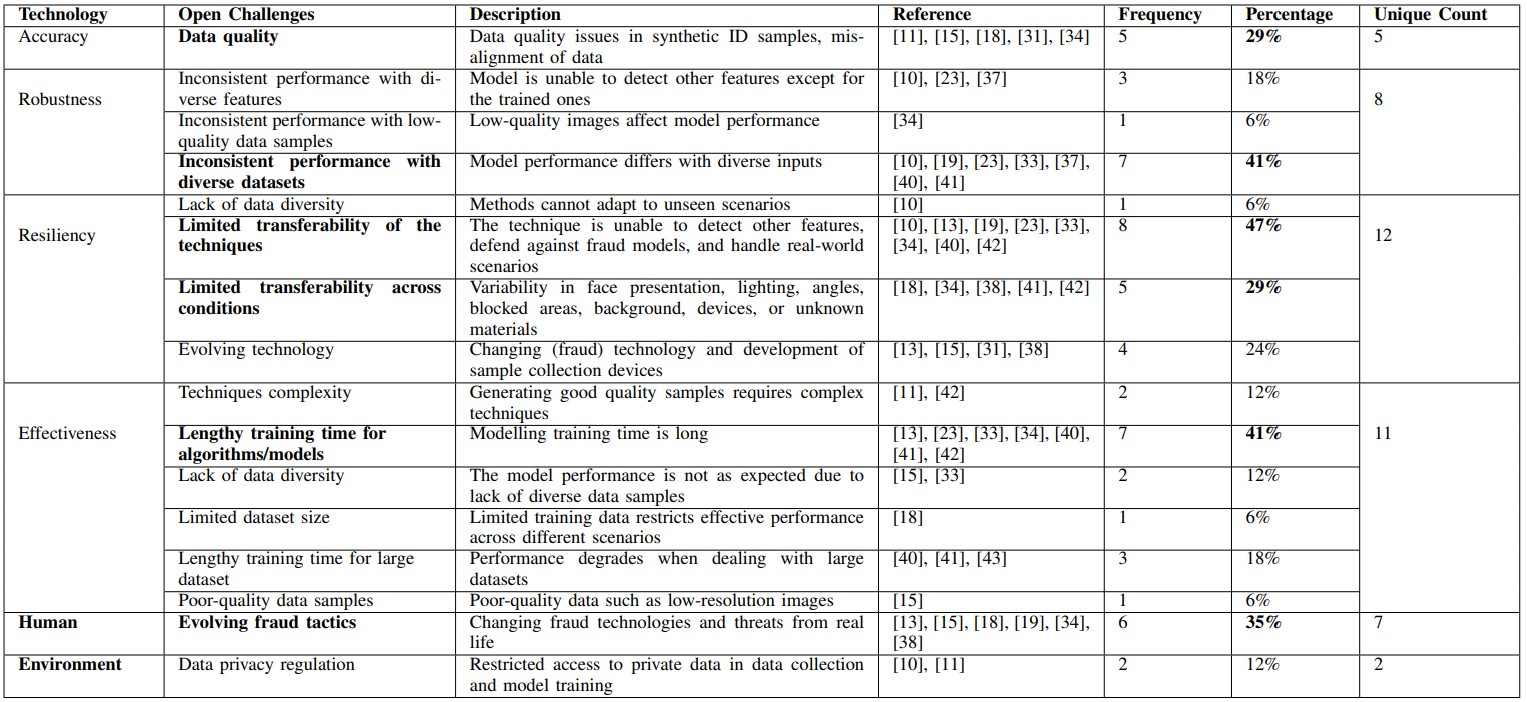} \\

    \end{tabular}

    \label{tab:9}
\end{table}

\subsection{AI-based IDF Prevention and Detection Method: Continuous Authentication}
This section focuses on continuous authentication, primarily implemented through user behaviour anomaly detection. The analysis of these methods is conducted within the framework of UEBA, which serves as the foundational operational principle to guide their implementation. Additionally, the open challenges were identified through frequency analysis.

\subsubsection{AI-based Continuous authentication method: user behaviour anomaly detection}

\begin{figure}
    \centering
    \includegraphics[width=0.5\linewidth]{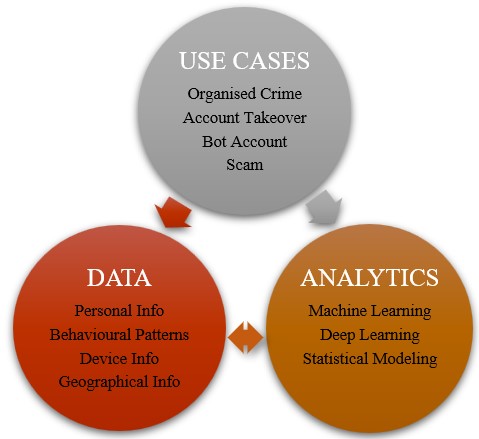}
    \caption{The Three Pillars of UEBA (Adapted from [92]}
    \label{fig:8}
\end{figure}

User behaviour anomaly detection utilises AI techniques, such as machine learning and deep learning, to identify deviations from normal behaviour in user profiles. In the reviewed articles, the analysis reveals the application of methods based on UEBA, a widely adopted approach in industrial practices [92]. These methods construct user profiles by analysing user behaviour patterns and their interactions with entities such as devices and applications [91]. This profile serves as a baseline for the users and is compared with real-time user data to identify potential anomalies [91-92]. Fig \ref{fig:8} shows the three pillars that underpin the functionality of the UEBA tool. 

The user behaviour anomaly detection methods are analysed using the dimensions of the UEBA framework. The analysis is organised into five key aspects: (a) core data for user profile creation, (b) profile modelling techniques, (c) analytics techniques, (d) use cases, and (e) open challenges. 

\paragraph{Core Data for User Profile Construction}

User profiles provide an important baseline for analysing user behaviour patterns and detecting anomalies [91]. Fig \ref{fig:7} illustrates the data and feature elements identified in the reviewed articles. These components, essential for user profile construction, are categorised as follows:
\begin{itemize}
    \item \textit{Personal information:} basic details such as name, date of birth, address, email address, phone, etc. [1], [2], [3], [12]. 
\item \textit{Behavioural Patterns:} the typical actions and interactions that users exhibit within a system. It includes these categories as in the followings:
   \begin{itemize}
      \item \textit{Device interaction:} user's interaction with such as touchscreen [30], zoom gestures [29], keystrokes [28], browser fingerprint [3] and sensor [39].
\item \textit{Login activity:} frequency, time, and location of logins [9].

   \item \textit{Social features:} user data linked to an account and its posts on an online social platform, such as friend counts, like counts, comments, etc. [16], [17], [21], [35], [9]. 
   \item \textit{Transaction history:} details of performed transactions, including amounts and recipients, etc., [4], [5], [6], [7], [9].
   \item \textit{Communication patterns:} the distinctive ways individuals express themselves through language, such as word choice, sentence structure, tone, and writing styles [8], [17], [20], [35],[16].
\end{itemize}
\item Device Information: details about devices used, such as IP addresses, MAC addresses, and device types [1], [3],[12].
\item Geographical information: regular physical locations from which the user accesses the system [12].
\end{itemize}
\begin{figure}[ht] 
    \raggedright
    
\includegraphics[scale=0.6]{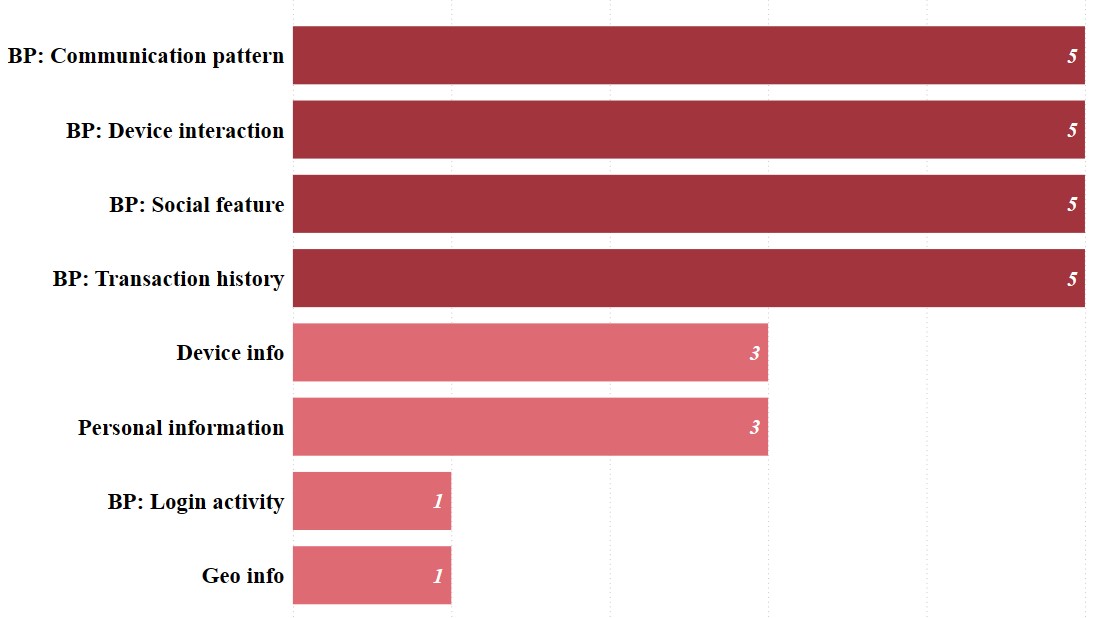}
\caption{Distribution of Attributes for Building User Profiles
}
\vspace{2mm}
\footnotesize
\textit{Note: Because some articles used multiple data types, the results reflect an aggregated analysis of the overall data. (Abbreviation refers to TABLE \ref{tab:8})}
    \label{fig:9}
\end{figure} 
Fig \ref{fig:9}
illustrates the distribution of attributes used to construct
user profiles in this review based on an aggregated
analysis. Communication patterns, device interaction,
social features, and transaction history are equally important attributes for building user profiles, with each
being utilised in five articles

\paragraph{User Profile Modelling Techniques}Due to differences in modelling purposes, the modelling techniques of user profiles can be broadly categorised into two types: historical patterns modelling and relational patterns modelling. Their workflows for anomaly detection differ slightly. 
\begin{itemize}
    \item  \textit{Historical patterns modelling:} this is the most common approach, utilised in 15 articles [4], [5], [6], [7], [8], [9], [16], [17], [20], [21], [28], [29], [30], [35], [39]. First, behaviour data of users and entities is collected from various data sources (e.g., logs, devices, application activities, etc.) and preprocessed. Next, feature extraction is performed. Due to the diversity of behavioural data, this step is particularly crucial. Feature engineering, a method used to extract more representative features from the data,  has become prevalent in user behaviour anomaly detection [5], [12], [9], [16], [17], [20], [21], [29], [30], [35]. It helps models focus on more relevant features and effectively represent user profiles. Subsequently, historical behaviour patterns are modelled. The model learns the normal behaviour of users as a baseline. When new user data is received, the model automatically identifies whether the new behaviour is expected or anomalous.
    \item  \textit{Relational patterns modelling:} This method first analyses labelled business data to identify fraudster behaviour patterns and then constructs these relational patterns using graph techniques [1], [2], [3], [12]. It highlights the interrelations among users, which support the mapping of networks that underpin organised crime.
    Therefore, the model detects anomalies by analysing whether new data points align with the characteristics of the identified anomaly group.
 \end{itemize}

\paragraph{Analytics Techniques}

\begin{figure}[ht] 
    \raggedright
    
\includegraphics[scale=0.6]{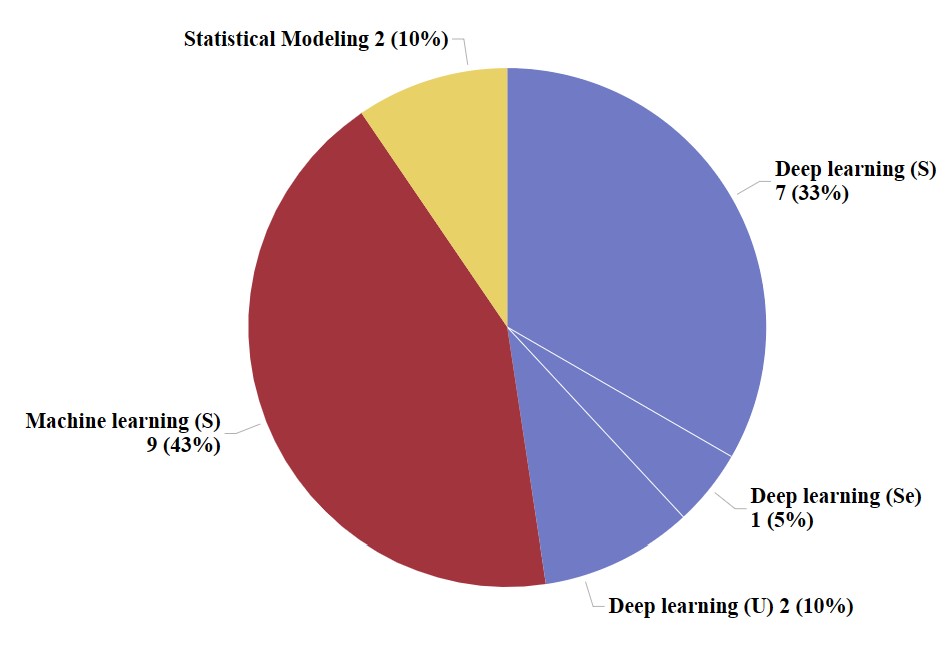}
\caption{Distribution of Analytics Techniques for User Behaviours Anomaly Detection
}
    \label{fig:10}
\end{figure}

This review identified analytics techniques, including machine learning, deep learning and statistical modelling, to identify and analyse abnormal behaviour patterns [92]. Fig\ref{fig:10} presents the distribution of analytics techniques in this review.

\begin{itemize}
    \item \textit{Deep learning:} 
The reviewed articles identify deep learning as the most extensively applied technique, referenced in 10 studies. Of these, seven use supervised learning [1], [5], [8], [20], [21], [28], [30], two use unsupervised learning [3], [12], and one employs a semi-supervised approach [2]. Except for [5], [21], and [30], the reviewed deep learning methods demonstrate significant potential in addressing complex tasks. These tasks include processing unstructured data such as graphs, text, and audio [1], [8], [20], [28]. These techniques also decrease reliance on human intervention such as data labelling and model optimisation, and improve the accuracy of predictions for previously unknown suspicious points [2], [3]. For instance, [2] employs a semi-supervised learning, training a Graph Neural Network (GNN) with labelled anomalous nodes. The GNN then infers additional suspicious nodes based on similarities within anomalous communities. Similarly, [3] employs unsupervised learning, such as clustering, DeepWalk, and probability-based models, to identify anomalous node groups and predict additional suspicious points.
    \item \textit{Machine learning:}Machine learning is the second most widely used technique, featured in 9 articles [4], [6], [7], [9], [16], [17], [21], [29], [30]. All machine learning methods employ supervised learning, which involves human input to label users as either legitimate or malicious. Support Vector Machine (SVM) is the most commonly used algorithm, known for its high accuracy in detecting anomalies [4], [6], [16], [17], [29]. Random Forest is the second most widely applied algorithm, also demonstrating strong performance [4], [16], [17], [29]. The reviewed articles suggest that machine learning models are particularly effective at classification tasks, with notable success in establishing behavioural baselines and detecting anomalies (Appendix \ref{appendix:B}).  
    \item \textit{Statistical modelling:} 
Two articles employ statistical modelling to detect anomalous users, through statistical methods and distribution-based analysis. [35] utilises the F-test to define a baseline, represented by a behaviour feature matrix derived from social features. The F-test is then applied to evaluate new user data and determine the presence of anomalous behaviour. In contrast, [39] adopts probabilistic modelling, constructing user profiles through the Gaussian Process Model. This method constructs a covariance matrix to predict the likelihood of new data points within the learned distribution. Statistical modelling effectively uncovers anomalies unnoticed by humans and performs well with limited samples and features [35], [39].
    
\end{itemize}     

\paragraph{Use Cases}
A use case represents a specific scenario or problem that a method is designed to address such as detecting fraud. 
User behaviour anomaly detection is applied to defend against various identity fraud cases. Among the reviewed articles, account takeovers are the most frequently addressed use case with 11 articles on this topic [4], [5], [6], [7], [9], [20], [28], [29], [30], [35], [39], followed by organised crime [1], [2], [3], [12] and bot accounts [16],[17],[21], while scam detection is the least discussed [8].

\paragraph{Open Challenges}

The open challenges in user behaviour anomaly detection reveal a complex interplay of technical and non-technical factors, with non-technical issues, such as changing user behaviour and data privacy regulations, being the most frequently highlighted. Key technical challenges include the limited diversity of datasets and the lack of transferability of techniques to real-world use cases. The following section elaborates on these challenges derived from TABLE \ref{tab:11}: 

\begin{itemize}
    \item Dynamic user behaviour, shaped by factors such as changing life events, priorities, and preferences, introduces significant variability that impacts the detection accuracy of methods [28], [29], [30]. [35], [39]. Consequently, anomalies in data do not necessarily indicate actual abnormal behaviour. As noted in [3], detected anomalies may include cases involving normal users. To mitigate this issue,  incorporating expert opinions and business-specific rules has been suggested in [3] and [7] to refine models and reduce false positives.
    \item Threshold misconfiguration is a frequently cited challenge affecting accuracy. While the rules outlined in [3] and [7] effectively distinguish normal users from anomalies, improper threshold settings can still lead to false negatives or false positives. Achieving an optimal balance in rule configuration remains a critical challenge, as it directly impacts detection performance and system reliability. 
    \item Restricted access to privacy data and user consent emerge as two key highlights identified in the review. Studies such as [3], [7], and [28] have raised concerns about accessing private user data for training purposes. Additionally, [5] and [9] discuss the impact of user consent policies on model performance and validation, particularly when users withdraw their data after training, posing challenges to both compliance and model reliability.
    
    \item Inconsistent performance across diverse datasets is a frequently observed issue, particularly in cases involving imbalanced datasets [29], [8]. This challenge is further compounded by the scarcity of labelled true fraud cases in user behaviour anomaly detection, which limits the model's ability to learn and generalise suspicious features due to insufficient data representation [8]. Additionally, studies such as [1], [20], and [35] identify challenges associated with real-time datasets and datasets tailored to varying use cases. For instance, while some methods effectively detect shared device relationships, they often lack sensitivity to subtle anomalies in individual behavioural characteristics. These findings highlight the difficulties in achieving consistent and reliable performance across diverse and dynamic datasets.
    
    \item Technique complexity is a significant factor impacting method effectiveness, particularly for those requiring intricate data processing and modelling [39], [20], [4], [6]. For instance, [39] describes a method that processes multiple sources of sensor data through unification, sequence conversion, and subsequent model training and classification. This multi-step pipeline is resource-intensive and poorly suited for large-scale datasets, creating significant barriers to real-time applications.

\end{itemize}
TABLE \ref{tab:10} summarises the articles on user behaviour anomaly detection. TABLE \ref{tab:12} summarises the three methods at each phase of the analysis, detailing the data type, application, applied context, and their advantages and disadvantages. Appendix \ref{appendix:B} summarises AI’s application and presents the performance metrics of the best models identified across all methods.

\begin{table}
    \centering
        \caption{User Behaviours Anomaly Detection Method's Article Summary}
    \begin{tabular}{c}
\includegraphics[width=0.97\textwidth]{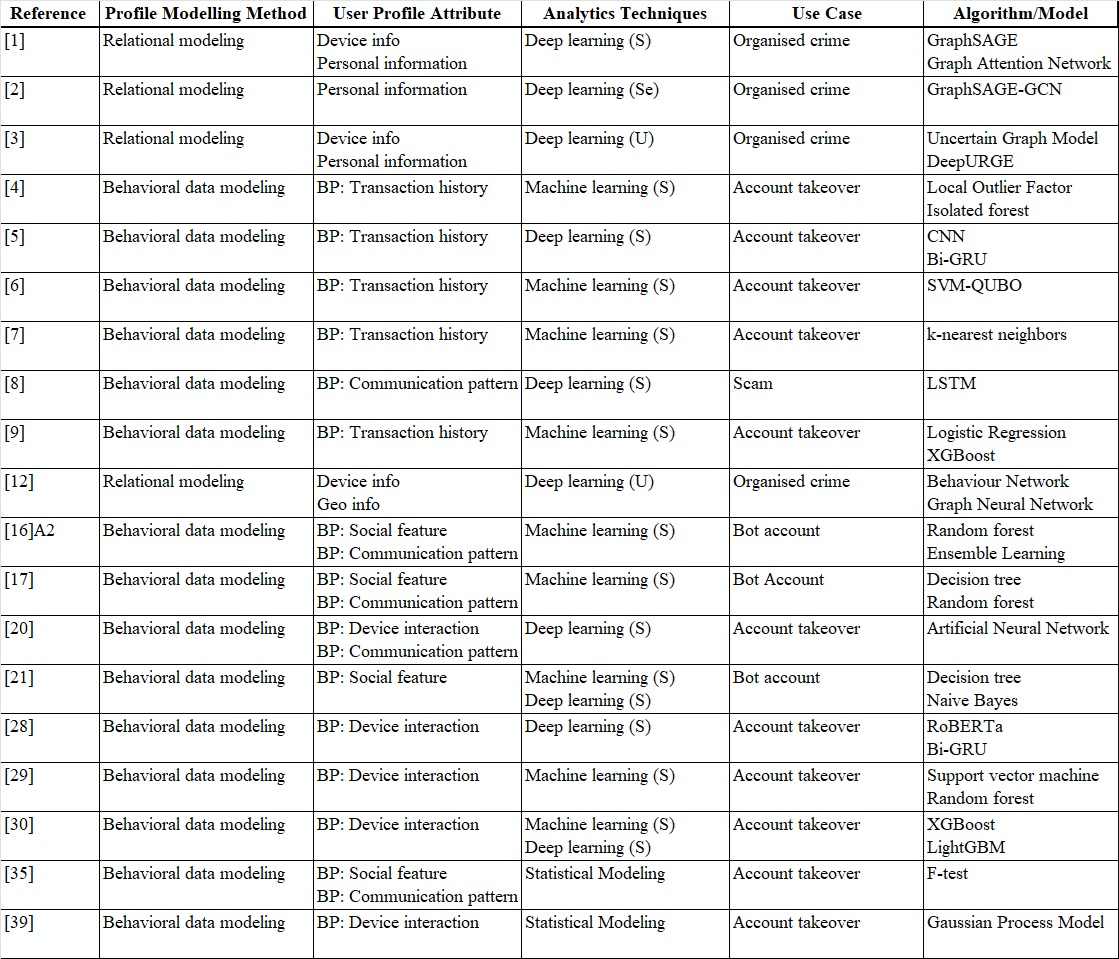} \\

    \end{tabular}

    \label{tab:10}
\end{table}

\begin{table}
    \centering
        \caption{Open Challenges in User Behaviour Anomaly Detection}
    \begin{tabular}{c}
\includegraphics[width=0.97\textwidth]{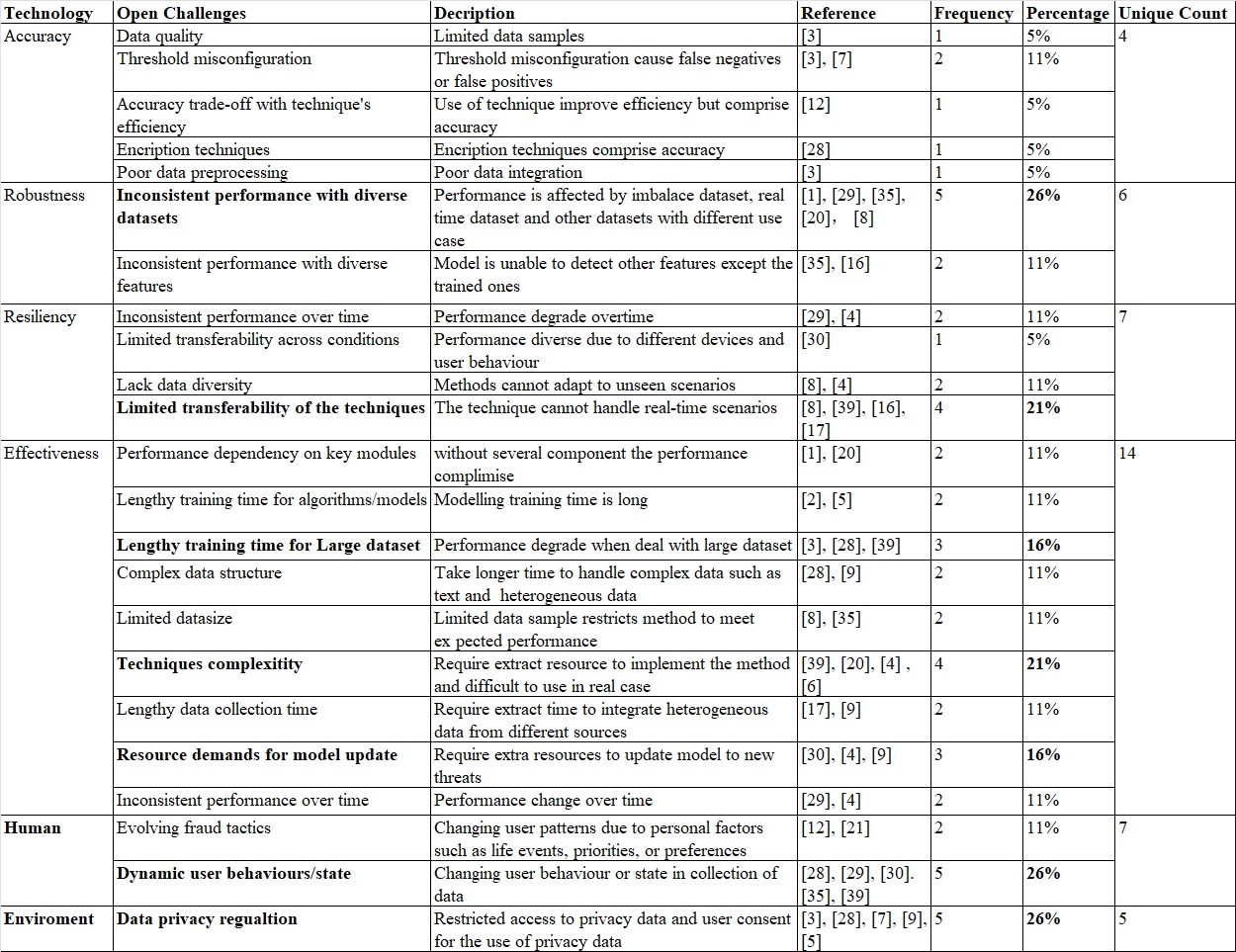} \\

    \end{tabular}

    \label{tab:12}
\end{table}

\begin{table}
    \centering
        \caption{Summary Of IDF Detection And Prevention Method}
    \begin{tabular}{c}
\includegraphics[width=0.97\textwidth]{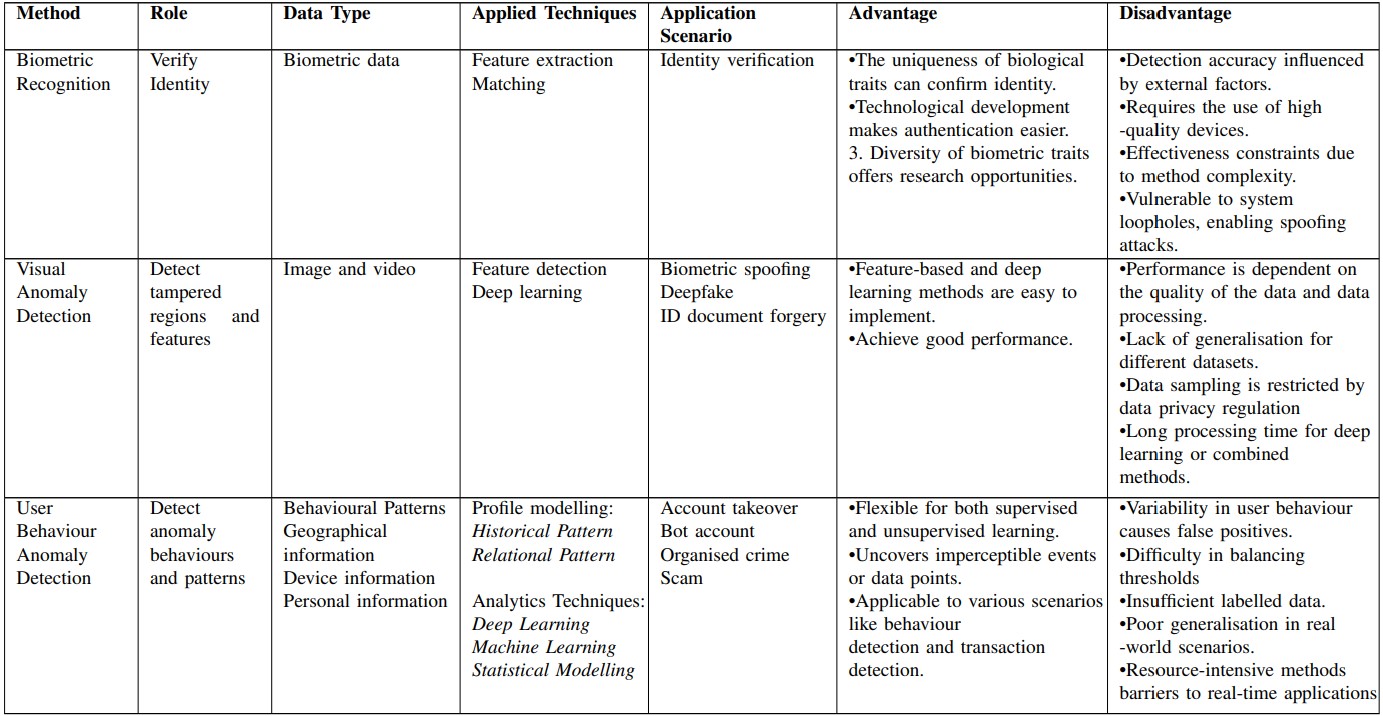} \\

    \end{tabular}

    \label{tab:11}
\end{table}

\section{Discussion}
This section addresses the research questions by analysing the findings and their broader implications. It explores how the results contribute to understanding the topic, suggests directions for future research and identifies the limitations of this review.

\subsection{Research Questions, Key Findings and Open Challenges}

This section addresses the research questions by summarising the key findings derived from the analysis. Each research question is discussed individually to ensure clarity and relevance.
\subsubsection{\textbf{RQ1: What are the current AI-based identity fraud detection methods}}

\begin{figure}
    \centering
    \includegraphics[width=0.9\linewidth]{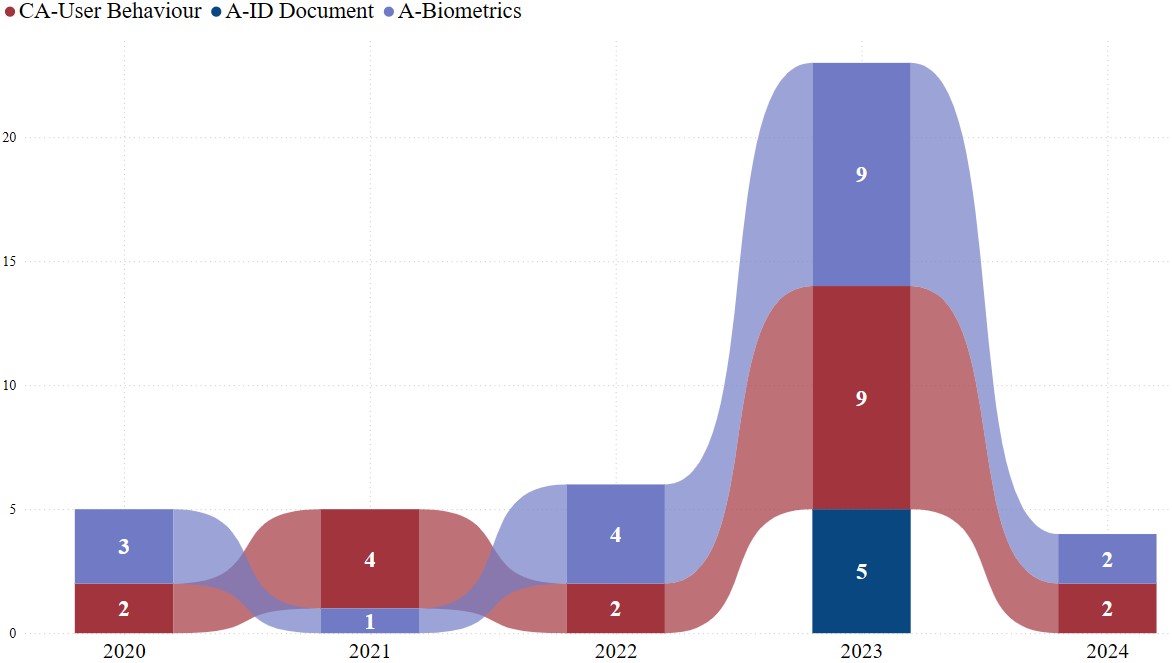}
    \caption{Trends in AI-Based Methods for Authentication and Continuous Authentication}
    \label{fig:11}
\footnotesize
\raggedright
\textit{\textbf{Note:} Abbreviations are defined as follows:}
\begin{enumerate}
    \item \textit{A: Authentication;}
    \item \textit{CA: Continuous Authentication;}
    \item \textit{A-Biometrics: Biometrics Authentication;}
    \item \textit{A-ID Document: ID Document Authentication.}
\end{enumerate}
\end{figure}

Given the impact of authentication and continuous authentication on combating identity fraud, this SLR carefully selected and reviewed 43 articles that applied AI-based methods to detect and prevent identity fraud in the two authentication processes. There is a noticeable increase in the number of articles that use AI-based methods to address identity fraud for the two authentication processes, as in Fig \ref{fig:11}. 

\begin{figure}[ht] 
    \raggedright
    
\includegraphics[scale=0.6]{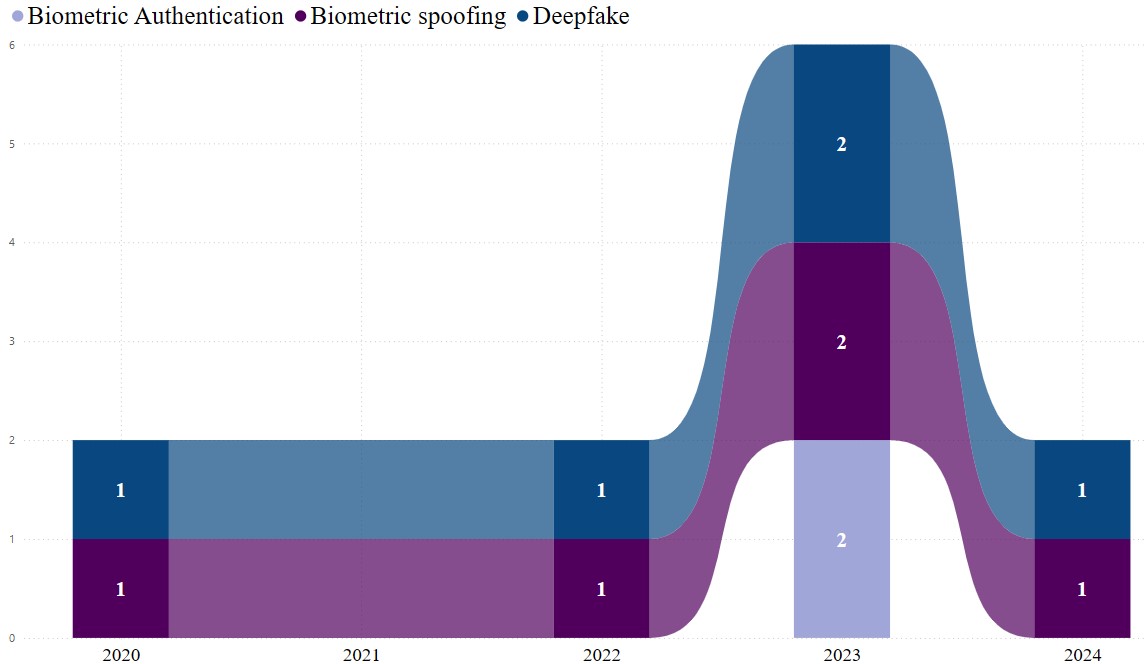}
    \caption{Trends in the Number of Articles on Face Recognition Over the Years
}
    \label{fig:12}
\end{figure}

In biometric recognition, facial recognition has emerged as the most discussed and extensively utilised biometric method in this field. Its application in secure access systems and mobile device authentication is driven by its ability to offer high accuracy and ease of use [27], [41].  The increasing deployment of facial recognition systems presents critical challenges, particularly in the areas of privacy and security. Among these, biometric spoofing and deepfake attacks are the most significant threats (Fig \ref{fig:12}). Simultaneously, the emergence of GAI technologies has further accelerated the evolution of fraud techniques [23], [37],[40]. This rapid advancement has likely driven intensified research efforts to address these risks. Moreover, it underlines the urgency of developing resilient solutions to ensure the trustworthiness and long-term sustainability of these technologies. Fig \ref{fig:12} shows the yearly trend in articles on face recognition, categorised by the research problem they addressed. 

Continuous authentication has become increasingly important after initial authentication to combat identity fraud on a broader scale[28] (See Fig \ref{fig:11}). Communication patterns have gained increasing recognition as a crucial component in user behaviour anomaly detection, now on par with conventional factors such as device interactions, social features, and transaction history for user profile creation (See Fig \ref{fig:9}). This growing prominence aligns with advancements in GAI, which have significantly improved their ability to interpret and analyse semantic content. Consequently, contextual data analysis is becoming a valuable approach for assessing user identity. For instance, [28] illustrated that unique writing styles, influenced by personal background and education, offer distinguishing characteristics that reinforce the utility of communication patterns in continuous authentication.

Deep learning has emerged as a dominant technology across various methods in identity fraud detection, including biometric recognition, visual anomaly detection and user behaviour anomaly detection.
This progress is particularly evident in the application of CNNs, which excel at detecting subtle visual irregularities [15],[18], [25], [32], 37], [40], and GNNs, which effectively model user relational profiles using graph-based representations[1-3], [12]. These developments highlight deep learning’s exceptional ability to process complex data modalities such as images, graphs, and text, and to automate complex detection tasks with minimal human intervention. Such capabilities position deep learning as a cornerstone for future innovations in identity fraud detection systems. However, this wide applicability is not without challenges. One significant limitation is the dependence on large-scale and diverse annotated datasets, particularly in areas like user behaviour analysis, where labelled data are scarce and expensive to obtain[10],[11]. Moreover, the high computational requirements of deep learning models often hinder their deployment in environments with limited resources such as mobile devices operating in real-time scenarios[13]. 

\subsubsection{\textbf{RQ2: What are the key open challenges in addressing identity fraud}}
The results of this research indicate that certain challenges are repeatedly mentioned 
in the literature. The findings emphasise the need to address these recurring challenges to advance research and improve practical implementations.

\begin{enumerate}
\item[a)]\textit{Challenge 1:Technical Limitations}

A significant challenge in this field lies in overcoming technical limitations that hinder the performance of current methods (See TABLE \ref{tab:13}).

\begin{itemize}

\item \textbf{Robustness:} The lack of diverse data samples is the primary technical challenge for IDF detection methods. These methods often struggle to detect inputs outside the learned patterns. 

\item \textbf{Resilience:} The limited transferability of techniques is the second major challenge. This suggests that while current AI methods perform well on familiar patterns, their performance significantly declines when applied to unfamiliar scenarios, such as dynamic or real-world environments. These challenges underscore the need for more robust and adaptable approaches to address the evolving nature of identity fraud.

\item \textbf{Accuracy:} Data quality is critical for the accuracy of IDF detection methods. Challenges arise from noisy samples and inadequate preprocessing techniques. Addressing these issues requires diverse data sampling strategies that capture real-world variations, such as environmental conditions and device characteristics. Metadata collected during sampling can support targeted corrections during preprocessing.

\item \textbf{Effectiveness:} Lengthy training times represent a major challenge in terms of method effectiveness, particularly in deep learning-based approaches. The computational demands of these models often result in extended training durations, which can hinder their scalability and applicability in real-time or resource-constrained environments. These limitations reinforce the need for more efficient model architectures and training algorithms to optimise performance.
\end{itemize}

\item[b)] \textit{Challenge 2:Contextual Variability}

Although significant research has been conducted on identity fraud, the understanding of its complex dynamics and mechanisms remains limited. One of the primary challenges is the continuous evolution of fraud tactics driven by advancing fraudulent technologies and emerging real-world threats, such as the rise of GAI. Business data from companies can assist in identifying suspicious patterns and labelling data for machine learning; however, this often occurs after fraud has already taken place. When new patterns emerge, models frequently struggle to make accurate predictions or adapt to novel situations.

Dynamic user behaviours further complicate the issue. As user behaviour constantly evolves, baseline patterns established by these methods may no longer hold true. Additionally, anomalies identified by detection systems do not always indicate actual fraud. Further analysis is often required to distinguish true fraud from benign outliers [3], [7].

\item[c)] \textit{Challenge 3:Data Privacy and Security Issues}

Data privacy and security remain critical in identity fraud detection. Data privacy regulation has restricted the use of privacy data, which is essential for good model performance.

Although data encryption techniques have been applied for years, encrypted data will comprise the model performance to find meaningful information from the data, especially in today’s context, where contextual data are used by an increasing number of researchers [28]. 

The challenge extends to protecting private data stored in centralised databases, a common practice among many organisations [9]. Centralised storage increases the risk of identity theft, especially when AI models rely on such data for training. Though federated Learning is effective in some cases, it's not always a viable solution [9]. This is particularly true for global issues like identity fraud, which spans across multiple industries and sectors rather than being confined to a single company. A single method may not address the broader, cross-industry scale of identity fraud.

\end{enumerate}

\subsection{Research Implications and Future Direction}
The challenges outlined in the previous section serve as a foundation for identifying potential research directions. By addressing these challenges, IDF detection can achieve significant advancements in the following areas.
\subsubsection{Insufficient Quality Data Samples and Lack of Data Diversity for Model Training}

The scarcity and homogeneity of data samples, driven by privacy regulations and limited access to real-world scenarios, significantly hinder effective model training. Although synthetic data generated using advanced techniques can increase sample sizes, it often falls short in representing diverse and unforeseen conditions. 

This gap highlights the need for more innovative methods to improve both the quality and diversity of data. Research should particularly focus on designing and validating synthetic data approaches that better mimic realistic variations, thus enhancing model robustness and adaptability to the dynamic and evolving nature of identity fraud.
\subsubsection{Limited understanding of identity fraud dynamics } Despite progress in understanding identity fraud, current methods rely heavily on post-incident data, limiting their ability to adapt to new and evolving fraud patterns. Future research should focus on developing innovative methods to detect and differentiate suspicious activities from normal behaviors, particularly in the absence of pre-existing fraud patterns or labeled data. Advancing anomaly detection techniques that do not rely on pre-labeled datasets is crucial for enabling models to adapt quickly and effectively to evolving fraud tactics. These efforts will facilitate the development of more robust and proactive fraud detection systems, capable of responding to the dynamic nature of identity fraud.
\subsubsection{Data Privacy and Security Challenges}
Ensuring data privacy and security remains a critical issue. Future discussions should address two key questions:

\begin{itemize}

    \item How can privacy-preserving methods be implemented during data processing without compromising data utility for analytical purposes?
    \item How can data lifecycle management be optimised to address identity fraud at a cross-industry scale while balancing privacy, security, and data utility?
\end{itemize}
Research should emphasise the development of frameworks and tools that uphold data security and compliance while supporting effective fraud detection efforts. These areas highlight the need for interdisciplinary collaboration to create innovative solutions that address the evolving challenges of identity fraud.

\subsubsection{Application of Multimodal Data in Identity Fraud Detection}

As AI’s ability to understand semantics improves, multimodal data have been increasingly utilised in identity fraud detection. Methods employing image data, behavioural data, and time-series data are already well-established; however, context-based detection techniques remain less commonly implemented. Future research should focus on developing effective methods to integrate context-based data, exploring strategies to preserve critical information for model learning, such as temporal or spatial relationships, while ensuring robust privacy measures to protect sensitive user data.

The findings of this SLR highlight the importance of adopting advanced AI-driven methods, such as anomaly detection and behavior modeling, to enhance fraud detection and prevention systems. Adaptive, scalable, and efficient models capable of handling dynamic user behaviours and evolving fraud tactics are essential for both industrial applications and academic advancements. Additionally, there is a pressing need to explore unsupervised learning and robust anomaly detection techniques that do not rely on pre-labelled data. Addressing challenges in data quality, contextual variability, and real-world applicability requires interdisciplinary approaches and stronger collaboration between academia and industry.

\subsection{Limitations}
Like any other SLR study, this research has limitations that should be considered when interpreting the findings. First, it only relies on four databases, which are limited in number, however they provide sufficient coverage of the existing literature. It is worth mentioning that the review was conducted within a specific time frame, and studies published after this period may not have been included, which potential overlook recent advancements. However, this SLR study provides a solid foundation for future studies by other researchers. 

Second, due to the scope of this research, a specific set of search strings was employed, which may have influenced the breadth of the results. While efforts were made to balance specificity and comprehensiveness, it is possible that some relevant studies were omitted due to variations in terminology or indexing practices. 

Furthermore, the review focuses on the domain of identity fraud detection and may have overlooked insights from adjacent fields such as cybersecurity which could provide valuable complementary perspectives. However, it was important to keep the focus on the identity fraud detection rather a broad SLR across cybersecurity. Here, it is important to mention that research team met regulatory to review the results and discuss any individual research or researcher bias with a view to maintain the study focus and improve the quality of results.   

Finally, the rapid pace of advancements in AI and related technologies poses a challenge in maintaining the relevance of the findings. While the review provides a comprehensive snapshot of the field, some trends or methods discussed may already be evolving, which requires future studies to continuously update the state of knowledge.

\section{Conclusion}
Identity fraud can have significant impacts on individuals and organisations, making it essential to understand and investigate methods for its early detection. AI has demonstrated its potential in various tasks, including computer vision, natural language processing, and graph processing, and our systematic literature review highlights its effectiveness in identity fraud detection. From our findings, anomaly detection, either on image or behaviour data, is the main method in identity fraud detection. Deep learning has higher capabilities in handling complex scenarios that require the input of various data forms, from images to structured data, while machine learning has higher computation efficiency. Privacy data concerns with the use of AI also require attention to be extracted while implementing identity fraud detection.   

Overall, our systematic review highlights the transformative role of AI in combating identity fraud and identifies key areas for future research. We also provide a detailed list of current gaps and unresolved issues to guide future research in this area. Specifically, the next steps include:
\begin{itemize}
    \item Advancing unsupervised learning techniques;
\item Addressing data quality and diversity issues;
\item Developing adaptable models for dynamic and evolving fraud patterns;
\item Optimising resource efficiency for scalable systems;
\item Mitigating privacy issues to guarantee secure and compliant handling of sensitive data.
\item Enhancing detection through multimodal data integration.
\end{itemize}

\appendix
\renewcommand{\thesection}{Appendix \Alph{section}}

\section{SCORING TABLE}
\label{appendix:A}
\begin{longtable}{|c|c|c|c|c|c|c|c|}
\hline
\textbf{Reference} & \textbf{Purpose} & \textbf{Relevance} & \textbf{Presentation} & \textbf{Evaluation} & \textbf{Finding} & \textbf{Direction} & \textbf{Score} \\ \hline
[1] & 1 & 1 & 1 & 1 & 1 & 1 & 6 \\ \hline
[2] & 1 & 1 & 1 & 1 & 1 & 1 & 6 \\ \hline
[3] & 1 & 1 & 1 & 1 & 1 & 1 & 6 \\ \hline
[4] & 1 & 1 & 1 & 1 & 0 & 0 & 4 \\ \hline
[5] & 1 & 1 & 1 & 1 & 0 & 0 & 4 \\ \hline
[6] & 1 & 1 & 0 & 1 & 1 & 0 & 4 \\ \hline
[7] & 1 & 1 & 0 & 1 & 0 & 1 & 4 \\ \hline
[8] & 1 & 1 & 1 & 1 & 1 & 1 & 6 \\ \hline
[9] & 1 & 1 & 1 & 1 & 0 & 0 & 4 \\ \hline
[10] & 1 & 1 & 1 & 0 & 1 & 1 & 5 \\ \hline
[11] & 1 & 1 & 1 & 1 & 1 & 1 & 6 \\ \hline
[12] & 1 & 1 & 1 & 1 & 1 & 0 & 5 \\ \hline
[13] & 1 & 1 & 1 & 1 & 0 & 1 & 5 \\ \hline
[14] & 1 & 1 & 1 & 1 & 0 & 1 & 5 \\ \hline
[15] & 1 & 1 & 1 & 1 & 0 & 0 & 4 \\ \hline
[16] & 1 & 1 & 0 & 1 & 0 & 1 & 4 \\ \hline
[17] & 1 & 1 & 1 & 1 & 0 & 1 & 5 \\ \hline
[18] & 1 & 1 & 0 & 1 & 0 & 1 & 4 \\ \hline
[19] & 1 & 1 & 1 & 1 & 0 & 0 & 4 \\ \hline
[20] & 1 & 1 & 1 & 1 & 1 & 1 & 6 \\ \hline
[21] & 1 & 1 & 1 & 1 & 0 & 0 & 4 \\ \hline
[22] & 1 & 1 & 1 & 1 & 1 & 1 & 6 \\ \hline
[23] & 1 & 1 & 1 & 1 & 1 & 1 & 6 \\ \hline
[24] & 1 & 1 & 1 & 1 & 0 & 1 & 5 \\ \hline
[25] & 1 & 1 & 1 & 1 & 0 & 0 & 4 \\ \hline
[26] & 1 & 1 & 1 & 1 & 0 & 0 & 4 \\ \hline
[27] & 1 & 1 & 1 & 1 & 1 & 1 & 6 \\ \hline
[28] & 1 & 1 & 1 & 1 & 1 & 1 & 6 \\ \hline
[29] & 1 & 1 & 1 & 1 & 1 & 1 & 6 \\ \hline
[30] & 1 & 1 & 1 & 1 & 1 & 1 & 6 \\ \hline
[31] & 1 & 1 & 1 & 1 & 1 & 1 & 6 \\ \hline
[32] & 1 & 1 & 0 & 1 & 1 & 0 & 4 \\ \hline
[33] & 1 & 1 & 1 & 1 & 0 & 0 & 4 \\ \hline
[34] & 1 & 1 & 1 & 1 & 0 & 1 & 5 \\ \hline
[35] & 1 & 1 & 1 & 1 & 1 & 1 & 6 \\ \hline
[36] & 1 & 1 & 1 & 1 & 1 & 1 & 6 \\ \hline
[37] & 1 & 1 & 1 & 1 & 1 & 1 & 6 \\ \hline
[38] & 1 & 1 & 1 & 1 & 0 & 1 & 5 \\ \hline
[39] & 1 & 1 & 1 & 1 & 1 & 1 & 6 \\ \hline
[40] & 1 & 1 & 1 & 1 & 1 & 1 & 6 \\ \hline
[41] & 1 & 1 & 1 & 1 & 0 & 1 & 5 \\ \hline
[42] & 1 & 1 & 1 & 1 & 1 & 1 & 6 \\ \hline
[43] & 1 & 1 & 0 & 1 & 0 & 0 & 3 \\ \hline
\end{longtable}
\clearpage

\section{Summary OF AI's Applications and Effectiveness}
\label{appendix:B}

\begin{table}[ht] 
    \centering
    \includegraphics[width=0.97\textwidth]{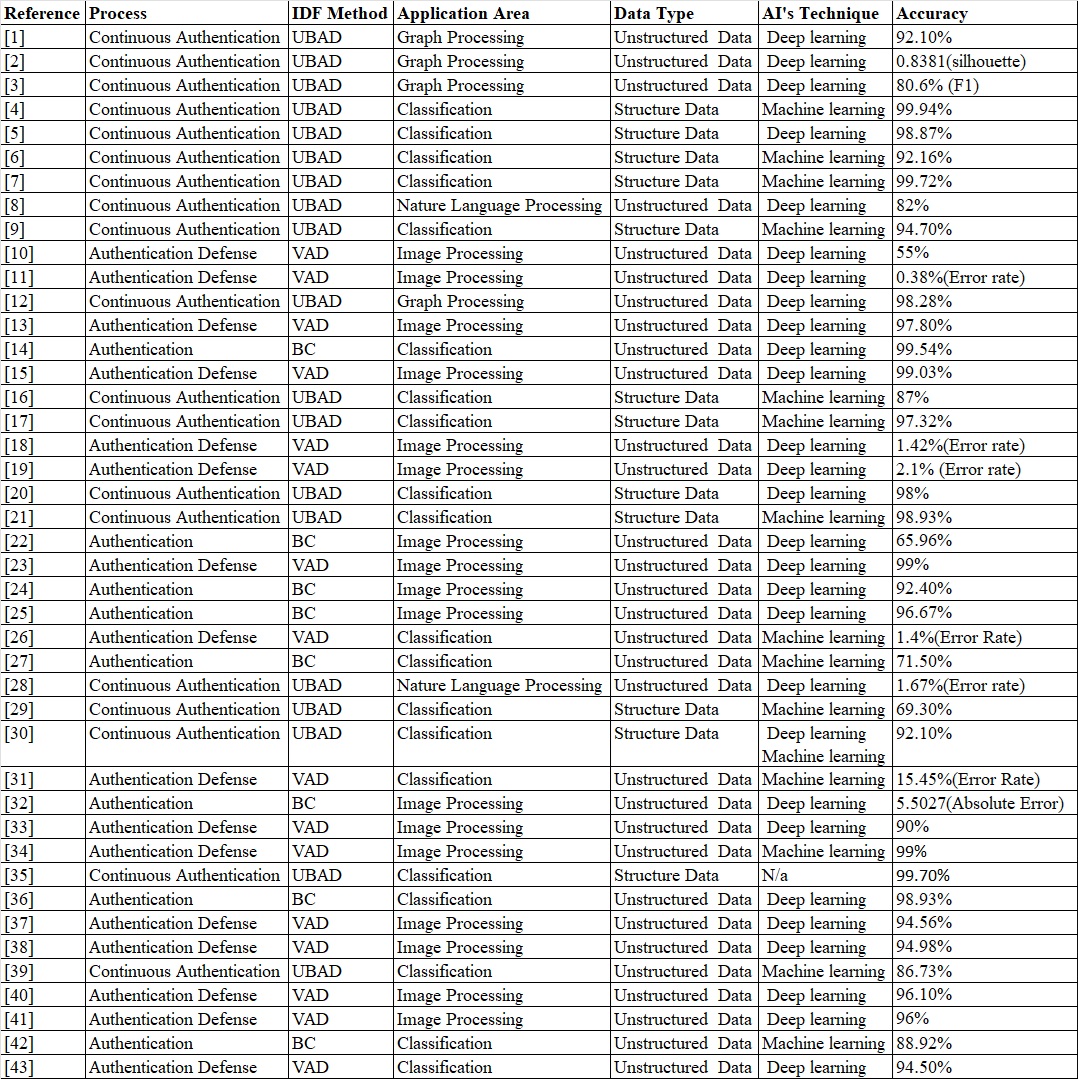} 
\end{table}

\bibliographystyle{unsrt}

\begin{thebibliography}{1}

\bibitem{Ref1}
\emph{Q. Ye et al.}, "Modeling Access Environment and Behavior Sequence for Financial Identity Theft Detection in E-Commerce Services," in \emph{2022 International Joint Conference on Neural Networks (IJCNN)}, 2022, pp. 1--8.

\bibitem{Ref2}
\emph{I. Khan et al.}, "Synthetic Identity Detection using Inductive Graph Convolutional Networks," in \emph{2023 10th International Conference on Computing for Sustainable Global Development (INDIACom)}, New Delhi, India, 2023, pp. 973--978, 2024.

\bibitem{Ref3}
\emph{Y.-W. Chang, H.-Y. Shih, and T.-N. Lin}, "AI-URG: Account Identity-Based Uncertain Graph Framework for Fraud Detection," \emph{IEEE Transactions on Computational Social Systems}, vol. 11, no. 3, pp. 3706--3728, 2024.

\bibitem{Ref4}
\emph{V. N. Dornadula and S. Geetha}, "Credit Card Fraud Detection using Machine Learning Algorithms," \emph{Procedia Computer Science}, vol. 165, pp. 631--641, 2019.

\bibitem{Ref5}
\emph{S. Kakkar et al.}, "Analysis of Discovering Fraud in Master Card Based on Bidirectional GRU and CNN Based Model," in \emph{2023 International Conference on Self Sustainable Artificial Intelligence Systems (ICSSAS)}, 2023, pp. 50--55.

\bibitem{Ref6}
\emph{T. Priyaradhikadevi et al.}, "Credit Card Fraud Detection Using Machine Learning Based on Support Vector Machine," in \emph{2023 Eighth International Conference on Science Technology Engineering and Mathematics (ICONSTEM)}, 2023, pp. 1--6.

\bibitem{Ref7}
\emph{A. Singhai, S. Aanjankumar, and S. Poonkuntran}, "A Novel Methodology for Credit Card Fraud Detection using KNN Dependent Machine Learning Methodology," in \emph{2023 2nd International Conference on Applied Artificial Intelligence and Computing (ICAAIC)}, 2023, pp. 878--884.

\bibitem{Ref8}
\emph{B. Hong, T. Connie, and M. K. Ong Goh}, "Scam Calls Detection Using Machine Learning Approaches," in \emph{2023 11th International Conference on Information and Communication Technology (ICoICT)}, 2023, pp. 442--447.

\bibitem{Ref9}
\emph{B. Lv et al.}, "Research on Modeling of E-banking Fraud Account Identification Based on Federated Learning," in \emph{2021 IEEE Intl Conf on Dependable, Autonomic and Secure Computing, Intl Conf on Pervasive Intelligence and Computing, Intl Conf on Cloud and Big Data Computing, Intl Conf on Cyber Science and Technology Congress (DASC/PiCom/CBDCom/CyberSciTech)}, 2021, pp. 611--618.

\bibitem{Ref10}
\emph{M. Al-Ghadi et al.}, "Guilloche Detection for ID Authentication: A Dataset and Baselines," in \emph{2023 IEEE 25th International Workshop on Multimedia Signal Processing (MMSP)}, 2023, pp. 1--6.

\bibitem{Ref11}
\emph{D. Benalcazar et al.}, "Synthetic ID Card Image Generation for Improving Presentation Attack Detection," \emph{IEEE Transactions on Information Forensics and Security}, vol. 18, pp. 1814--1824, 2023.

\bibitem{Ref12}
\emph{S. Hu et al.}, "Turbo: Fraud Detection in Deposit-free Leasing Service via Real-Time Behavior Network Mining," in \emph{2021 IEEE 37th International Conference on Data Engineering (ICDE)}, 2021, pp. 2583--2594.

\bibitem{Ref13}
\emph{J. Dong et al.}, "Restricted Black-Box Adversarial Attack Against DeepFake Face Swapping," \emph{IEEE Transactions on Information Forensics and Security}, vol. 18, pp. 2596--2608, 2023.

\bibitem{Ref14}
\emph{M. S. Islam and I. Elwarfalli}, "Deep Learning-Powered ECG-Based Biometric Authentication," in \emph{2023 International Conference on Next-Generation Computing, IoT and Machine Learning (NCIM)}, 2023, pp. 1--6.

\bibitem{Ref15}
\emph{G. Mazaheri and A. K. Roy-Chowdhury}, "Detection and Localization of Facial Expression Manipulations," in \emph{2022 IEEE/CVF Winter Conference on Applications of Computer Vision (WACV)}, 2022, pp. 2773--2783.

\bibitem{Ref16}
\emph{K. Umbrani et al.}, "Fake Profile Detection Using Machine Learning," in \emph{2024 ASU International Conference in Emerging Technologies for Sustainability and Intelligent Systems (ICETSIS)}, 2024, pp. 966--973.

\bibitem{Ref17}
\emph{A. Aguilera et al.}, "CrediBot: Applying Bot Detection for Credibility Analysis on Twitter," \emph{IEEE Access}, vol. 11, pp. 108365--108385, 2023.

\bibitem{Ref18}
\emph{V. Sharma et al.}, "U-Architecture for Face Recognition to Prevent Cyber and Spoofing Attacks in IoT," in \emph{2023 International Conference on Data Science and Network Security (ICDSNS)}, 2023, pp. 1--8.

\bibitem{Ref19}
\emph{Q. Guo et al.}, "Patch-Swap Based Approach for Face Anti-Spoofing Enhancement," in \emph{2023 IEEE Region 10 Symposium (TENSYMP)}, 2023, pp. 1--5.

\bibitem{Ref20}
\emph{F. Safara et al.}, "An Author Gender Detection Method Using Whale Optimization Algorithm and Artificial Neural Network," \emph{IEEE Access}, vol. 8, pp. 48428--48437, 2020.

\bibitem{Ref21}
\emph{J. Gangan et al.}, "Detection of Fake Twitter Accounts Using Ensemble Learning Model," in \emph{2023 7th International Conference On Computing, Communication, Control And Automation (ICCUBEA)}, 2023, pp. 1--6.

\bibitem{Ref22}
\emph{M. Hu and Y.-H. Hu}, "Machine Learning With Adaptive Image Colorization for Improving Facial Recognition," in \emph{2023 Congress in Computer Science, Computer Engineering, and Applied Computing (CSCE)}, 2023, pp. 2020--2024.

\bibitem{Ref23}
\emph{S. Agarwal et al.}, "Detecting Deep-Fake Videos from Appearance and Behavior," in \emph{2020 IEEE International Workshop on Information Forensics and Security (WIFS)}, 2020, pp. 1--6.

\bibitem{Ref24}
\emph{C. A. Krishna and R. Bhuvaneswari}, "Offline Signature Forgery Detection using Multi-Layer Perceptron," in \emph{2023 3rd Asian Conference on Innovation in Technology (ASIANCON)}, 2023, pp. 1--4.

\bibitem{Ref25}
\emph{B. Vinod and A. M. Senthil Kumar}, "Handwritten Signature Identification and Fraud Detection using Deep Learning and Computer Vision," in \emph{2023 International Conference on Sustainable Computing and Data Communication Systems (ICSCDS)}, 2023, pp. 196--200.

\bibitem{Ref26}
\emph{I. Avcibas}, "Morphed Face Detection With Wavelet-Based Co-Occurrence Matrices," \emph{IEEE Signal Processing Letters}, vol. 31, pp. 1344--1348, 2024.

\bibitem{Ref27}
\emph{A. Claveria et al.}, "Effects of Pressure on Acoustic Hand Biometric Authentication," in \emph{2021 IEEE MIT Undergraduate Research Technology Conference (URTC)}, 2021, pp. 1--5.

\bibitem{Ref28}
\emph{H. Yang et al.}, "CKDAN: Content and keystroke dual attention networks with pre-trained models for continuous authentication," \emph{Computers and Security}, vol. 128, 2023.

\bibitem{Ref29}
\emph{L. Wang et al.}, "The effectiveness of zoom touchscreen gestures for authentication and identification and its changes over time," \emph{Computers and Security}, vol. 111, 2021.

\bibitem{Ref30}
\emph{M. Levi et al.}, "Behavioral embedding for continuous user verification in global settings," \emph{Computers and Security}, vol. 119, 2022.

\bibitem{Ref31}
\emph{J. M. S. S. Thorpe}, "Classifying recaptured identity documents using the biomedical Meijering and Sato algorithms," 2023.

\bibitem{Ref32}
\emph{Z. N. Izdihar et al.}, "Age Identification Through Facial Images Using Gabor Filter and Convolutional Neural Network (CNN)."

\bibitem{Ref33}
\emph{A. K. Jaiswal and R. Srivastava}, "Fake region identification in an image using deep learning segmentation model," \emph{Multimedia Tools and Applications}, vol. 82, no. 25, pp. 38901--38921, 2023.

\bibitem{Ref34}
\emph{M.-N. Chapel, M. Al-Ghadi, and J.-C. Burie}, "Authentication of Holograms with Mixed Patterns by Direct LBP Comparison," in \emph{2023 IEEE 25th International Workshop on Multimedia Signal Processing (MMSP)}, 2023, pp. 1--6.

\bibitem{Ref35}
\emph{S. Hanok and S. Shankaraiah}, "In Loco Identity Fraud Detection Model using Statistical Analysis for Social Networking Sites: A Case Study with Facebook," \emph{The International Arab Journal of Information Technology}, vol. 20, no. 2, 2023.

\bibitem{Ref36}
\emph{R. Srivastva, Y. N. Singh, and A. Singh}, "Statistical independence of ECG for biometric authentication," \emph{Pattern Recognition}, vol. 127, p. 108640, 2022.

\bibitem{Ref37}
\emph{J. K and M. A}, "Safeguarding media integrity: A hybrid optimized deep feature fusion based deepfake detection in videos," \emph{Computers and Security}, vol. 142, 2024.

\bibitem{Ref38}
\emph{A. Trabelsi, M. Pic, and J.-L. Dugelay}, "Recapture Detection to Fight Deep Identity Theft," in \emph{Proceedings of the 2022 4th International Conference on Video, Signal and Image Processing}, 2022, pp. 37--42.

\bibitem{Ref39}
\emph{P. Zhao and M. Wang}, "Mobile behavior trusted certification based on multivariate behavior sequences," \emph{Neurocomputing}, vol. 419, pp. 203--214, 2021.

\bibitem{Ref40}
\emph{Y. Cao et al.}, "Three-classification face manipulation detection using attention-based feature decomposition," \emph{Computers and Security}, vol. 125, 2023.

\bibitem{Ref41}
\emph{D. M. Uliyan, S. Sadeghi, and H. A. Jalab}, "Anti-spoofing method for fingerprint recognition using patch based deep learning machine," \emph{Engineering Science and Technology, an International Journal}, vol. 23, no. 2, pp. 264--273, 2020.

\bibitem{Ref42}
\emph{F. A. Pujol et al.}, "Entropy-Based Face Recognition and Spoof Detection for Security Applications," \emph{Sustainability}, vol. 12, no. 1, 2019.

\bibitem{Ref43}
\emph{K. Nugroho and E. Winarno}, "Spoofing Detection of Fake Speech Using Deep Neural Network Algorithm," in \emph{2022 International Seminar on Application for Technology of Information and Communication (iSemantic)}, 2022, pp. 56--60.

\bibitem{Ref44}
\emph{S. R. Ahmed}, "Preventing Identity Crime: identity theft and identity fraud: An Identity Crime Model and Legislative Analysis with Recommendations for Preventing Identity Crime." BRILL, 2020.

\bibitem{Ref45}
\emph{K. George}, "Hong Kong CFO targeted in deepfake scam," \emph{CNN}, Feb. 4, 2024. [Online]. Available: https://edition.cnn.com/2024/02/04/asia/deepfake-cfo-scam-hong-kong-intl-hnk/index.html.

\bibitem{Ref46}
\emph{I. Vel\'asquez, A. Caro, and A. Rodr\'iguez}, "Authentication schemes and methods: A systematic literature review," \emph{Information and Software Technology}, vol. 94, pp. 30--37, Sep. 2017, doi: 10.1016/j.infsof.2017.09.012.

\bibitem{Ref47}
\emph{S. R. Kodituwakku}, "Biometric authentication: A review," \emph{International Journal of Trend in Research and Development}, vol. 2, no. 4, pp. 113--123, 2015.

\bibitem{Ref48}
\emph{A. F. Baig and S. Eskeland}, "Security, Privacy, and Usability in Continuous Authentication: a survey," \emph{Sensors}, vol. 21, no. 17, p. 5967, Sep. 2021, doi: 10.3390/s21175967.

\bibitem{Ref49}
\emph{S. Russell and P. Norvig}, \emph{Artificial Intelligence: A Modern Approach, Global Edition}. Pearson Higher Ed, 2021.

\bibitem{Ref50}
Art. 4 GDPR -- Definitions - \emph{General Data Protection Regulation (GDPR)}, \emph{General Data Protection Regulation (GDPR)}, Mar. 29, 2018. [Online]. Available: https://gdpr-info.eu/art-4-gdpr/.

\bibitem{Ref51}
\emph{B. Kitchenham, O. P. Brereton, D. Budgen, M. Turner, J. Bailey, and S. Linkman}, "Systematic literature reviews in software engineering -- A systematic literature review," \emph{Information and Software Technology}, vol. 51, no. 1, pp. 7--15, Nov. 2008, doi: 10.1016/j.infsof.2008.09.009.

\bibitem{Ref52}
\emph{W. Knight}, "OpenAI\'s new language generator GPT-3 is shockingly good -- and completely mindless," \emph{MIT Technology Review}, Jul. 20, 2020. [Online]. Available: https://www.technologyreview.com/2020/07/20/1005454/openai-machine-learning-language-generator-gpt-3-nlp/.

\bibitem{Ref53}
\emph{A. K. Jain and A. Kumar}, "Biometric Recognition: An Overview," in \emph{The International Library of Ethics, Law and Technology}, 2012, pp. 49--79. doi: 10.1007/978-94-007-3892-8\_3.



\bibitem{Ref54}
\emph{J. Yang, R. Xu, Z. Qi, and Y. Shi}, "Visual Anomaly Detection for Images: A Systematic survey," \emph{Procedia Computer Science}, vol. 199, pp. 471--478, Jan. 2022, doi: 10.1016/j.procs.2022.01.057.

\bibitem{Ref55}
\emph{Z. Iqbal, M. A. Khan, M. Sharif, J. H. Shah, M. H. U. Rehman, and K. Javed}, "An automated detection and classification of citrus plant diseases using image processing techniques: A review," \emph{Computers and Electronics in Agriculture}, vol. 153, pp. 12--32, Aug. 2018, doi: 10.1016/j.compag.2018.07.032.

\bibitem{Ref56}
Office of the Australian Information Commissioner (OAIC), \emph{Notifiable Data Breaches Report: January--June 2020}, 2020. [Online]. Available: https://www.oaic.gov.au/privacy/notifiable-data-breaches/notifiable-data-breaches-publications/notifiable-data-breaches-report-januaryjune-2020.

\bibitem{Ref57}
\emph{Office of the Australian Information Commissioner (OAIC)}, "Notifiable Data Breaches Report: July–December 2020," 2020. [Online]. Available: \url{https://www.oaic.gov.au/privacy/notifiable-data-breaches/notifiable-data-breaches-publications/notifiable-data-breaches-report-julydecember-2020}.

\bibitem{Ref58}
\emph{Office of the Australian Information Commissioner (OAIC)}, "Notifiable Data Breaches Report: January–June 2021," 2021. [Online]. Available: \url{https://www.oaic.gov.au/privacy/notifiable-data-breaches/notifiable-data-breaches-publications/notifiable-data-breaches-report-januaryjune-2021}.

\bibitem{Ref59}
\emph{Office of the Australian Information Commissioner (OAIC)}, "Notifiable Data Breaches Report: July–December 2021," 2021. [Online]. Available: \url{https://www.oaic.gov.au/privacy/notifiable-data-breaches/notifiable-data-breaches-publications/notifiable-data-breaches-report-july-to-december-2021}.

\bibitem{Ref60}
\emph{Office of the Australian Information Commissioner (OAIC)}, "Notifiable Data Breaches Report: January–June 2022," 2022. [Online]. Available: \url{https://www.oaic.gov.au/privacy/notifiable-data-breaches/notifiable-data-breaches-publications/notifiable-data-breaches-report-january-to-june-2022}.

\bibitem{Ref61}
\emph{Office of the Australian Information Commissioner (OAIC)}, "Notifiable Data Breaches Report: July–December 2022," 2022. [Online]. Available: \url{https://www.oaic.gov.au/privacy/notifiable-data-breaches/notifiable-data-breaches-publications/notifiable-data-breaches-report-july-to-december-2022}.

\bibitem{Ref62}
\emph{Office of the Australian Information Commissioner (OAIC)}, "Notifiable Data Breaches Report: January–June 2023," 2023. [Online]. Available: \url{https://www.oaic.gov.au/privacy/notifiable-data-breaches/notifiable-data-breaches-publications/notifiable-data-breaches-report-january-to-june-2023}.

\bibitem{Ref63}
\emph{Office of the Australian Information Commissioner (OAIC)}, "Notifiable Data Breaches Report: July–December 2023," 2023. [Online]. Available: \url{https://www.oaic.gov.au/privacy/notifiable-data-breaches/notifiable-data-breaches-publications/notifiable-data-breaches-report-july-to-december-2023}.

\bibitem{Ref64}
\emph{Office of the Australian Information Commissioner (OAIC)}, "Notifiable Data Breaches Report: January–June 2024," 2024. [Online]. Available: \url{https://www.oaic.gov.au/privacy/notifiable-data-breaches/notifiable-data-breaches-publications/notifiable-data-breaches-report-january-to-june-2024}.

\bibitem{Ref65}
G. Heusch, A. George, D. Geissbuhler, Z. Mostaani, and S. Marcel, "Deep models and shortwave infrared information to detect face presentation attacks," \emph{IEEE Transactions on Biometrics Behavior and Identity Science}, vol. 2, no. 4, pp. 399–409, Jul. 2020, doi: \url{10.1109/tbiom.2020.3010312}.

\bibitem{Ref66}
D. Rountree, "Federated Identity Primer," Chapter 1: "Introduction to Identity," Elsevier eBooks, pp. 1–11, 2013. [Online]. Available: \url{https://doi.org/10.1016/b978-0-12-407189-6.00001-7}. [Accessed: Mar. 12, 2024].

\bibitem{Ref67}
A. I. Segovia Domingo and Á. Martín Enríquez, "Digital Identity: the current state of affairs," 2018. [Online]. Available: \url{https://www.bbvaresearch.com/wp-content/uploads/2018/02/Digital-Identity_the-current-state-of-affairs.pdf}. [Accessed: Mar. 12, 2024].

\bibitem{Ref68}
ISO/IEC, "ISO/IEC 24760-1:2011: Information technology — Security techniques — A framework for identity management — Part 1: Terminology and concepts," International Organization for Standardization, 2011. [Online]. Available: \url{https://www.iso.org/standard/57914.html}. [Accessed: Mar. 12, 2024].

\bibitem{Ref69}
\emph{United Nations Office on Drugs and Crime}, "United Nations Handbook on Identity-related Crime," United Nations Office on Drugs and Crime, 2011. [Online]. Available: \url{https://www.unodc.org/documents/treaties/UNCAC/Publications/Handbook_on_ID_Crime/10-57802_ebooke.pdf}. [Accessed: Mar. 18, 2024].

\bibitem{Ref70}
Y. Irvin-Erickson, "Identity fraud victimization: a critical review of the literature of the past two decades," \emph{Crime Science}, vol. 13, no. 1, Feb. 2024, doi: \url{10.1186/s40163-024-00202-0}.

\bibitem{Ref71}
M. M. McNally and G. R. Newman, Eds., \emph{Perspectives on Identity Theft}. Monsey, N.Y.: Criminal Justice Press, 2008.

\bibitem{Ref72}
L. P. Leonardo Cavaliere \emph{et al.}, "The Impact of Internet Fraud on Financial Performance of Banks," \emph{Turkish Online Journal of Qualitative Inquiry}, vol. 12, no. 6, 2021.

\bibitem{Ref73}
M. Button and C. Cross, "Technology and Fraud: The ‘Fraudogenic’ consequences of the Internet revolution," in \emph{The Routledge Handbook of Technology, Crime and Justice}, Routledge, 2017, pp. 78–95.

\bibitem{Ref74}
G. Norris, A. Brookes, and D. Dowell, "The psychology of internet fraud victimisation: A systematic review," \emph{Journal of Police and Criminal Psychology}, vol. 34, pp. 231–245, 2019.

\bibitem{Ref75}
D. Modic and R. Anderson, "It's all over but the crying: The emotional and financial impact of internet fraud," \emph{IEEE Security and Privacy}, vol. 13, no. 5, pp. 99–103, 2015.

\bibitem{Ref76}
M. Button, C. Lewis, and J. Tapley, "Not a victimless crime: The impact of fraud on individual victims and their families," \emph{Security Journal}, vol. 27, pp. 36–54, 2014.

\bibitem{Ref77}
M. Button and C. Cross, "Technology and Fraud: The ‘Fraudogenic’ consequences of the Internet revolution," in \emph{The Routledge Handbook of Technology, Crime and Justice}, Routledge, 2017, pp. 78–95.

\bibitem{Ref78}
B. Green, S. Gies, A. Bobnis, N. L. Piquero, A. R. Piquero, and E. Velasquez, "Exploring Identity-Based Crime Victimizations: Assessing Threats and Victim Services among a Sample of Professionals," \emph{Deviant Behavior}, vol. 42, no. 9, pp. 1086–1099, Jan. 2020, doi: \url{10.1080/01639625.2020.1720938}.

\bibitem{Ref79}
\emph{Criminal Code 2002 | Acts}, ACT Legislation Register. [Online]. Available: \url{https://www.legislation.act.gov.au/a/2002-51/}.

\bibitem{Ref80}
T. F. Blauth, O. J. Gstrein, and A. Zwitter, "Artificial Intelligence Crime: An overview of Malicious use and abuse of AI," \emph{IEEE Access}, vol. 10, pp. 77110–77122, Jan. 2022, doi: \url{10.1109/access.2022.3191790}.

\bibitem{Ref81}
\emph{FedPayments Improvement}, "Toolkit Module 1: Synthetic Identity Fraud: The Basics," Feb. 17, 2022. [Online]. Available: \url{https://fedpaymentsimprovement.org/synthetic-identity-fraud-mitigation-toolkit/synthetic-identity-fraud-basics/}.

\bibitem{Ref82}
\emph{FedPayments Improvement}, "Toolkit Module 2: How Synthetic Identities are used," Feb. 17, 2022. [Online]. Available: \url{https://fedpaymentsimprovement.org/synthetic-identity-fraud-mitigation-toolkit/how-synthetic-identities-are-used/}.

\bibitem{Ref83}
U. Shafique\emph{et al.}, "Modern Authentication techniques in smart Phones: security and Usability perspective," \emph{International Journal of Advanced Computer Science and Applications}, vol. 8, no. 1, Jan. 2017, doi: \url{10.14569/ijacsa.2017.080142}.

\bibitem{Ref84}
D. D. Luxton, R. A. Kayl, and M. C. Mishkind, "MHealth Data Security: The need for HIPAA-Compliant Standardization," \emph{Telemedicine Journal and e-Health}, vol. 18, no. 4, pp. 284–288, Mar. 2012, doi: \url{10.1089/tmj.2011.0180}.

\bibitem{Ref85}
S. Lalchand, V. Srinivas, and J. Gregorie, "Using biometrics to fight back against rising synthetic identity fraud," \emph{Deloitte Insights}, Nov. 23, 2023. [Online]. Available: \url{https://www2.deloitte.com/us/en/insights/industry/financial-services/financial-services-industry-predictions/2023/financial-institutions-synthetic-identity-fraud.html}.

\bibitem{Ref86}
M. Masood, M. Nawaz, K. M. Malik, A. Javed, A. Irtaza, and H. Malik, "Deepfakes generation and detection: state-of-the-art, open challenges, countermeasures, and way forward," \emph{Applied Intelligence}, vol. 53, no. 4, pp. 3974–4026, Jun. 2022, doi: \url{10.1007/s10489-022-03766-z}.

\bibitem{Ref87}
J. Galbally, S. Marcel, and J. Fierrez, "Biometric Antispoofing Methods: A survey in face recognition," \emph{IEEE Access}, vol. 2, pp. 1530–1552, Jan. 2014, doi: \url{10.1109/access.2014.2381273}.

\bibitem{Ref88}
S. Ayeswarya and J. Norman, "A survey on different continuous authentication systems," \emph{International Journal of Biometrics}, vol. 11, no. 1, p. 67, Dec. 2018, doi: \url{10.1504/ijbm.2019.096574}.

\bibitem{Ref89}
A. Q. Gill, \emph{The Gill Framework: Adaptive Enterprise Architecture Toolkit}. CreateSpace Independent Publishing Platform, 2012.

\bibitem{Ref90}
V. Kharchenko, H. Fesenko, and O. Illiashenko, "Quality Models for Artificial Intelligence Systems: Characteristic-Based Approach, Development and Application," \emph{Sensors}, vol. 22, no. 13, p. 4865, Jun. 2022, doi: \url{10.3390/s22134865}.


\bibitem{ref91}
A. Litan, \emph{Market Guide for User and Entity Behavior Analytics}, ID G00276088, archived, Sep. 22, 2015. [Online]. Available: \url{https://www.gartner.com/document/3134524}. [Accessed: Nov. 22, 2024].

\bibitem{ref92}
G. Sadowski, J. Care, N. MacDonald, and H. Teixeira, \emph{Market Guide for User and Entity Behavior Analytics}, ID G00361156, archived, May 21, 2019. [Online]. Available: \url{https://www.gartner.com/document/3134524}. [Accessed: Nov. 22, 2024].

\bibitem{ref93}
A. Ali \emph{et al.}, “Financial Fraud Detection Based on Machine Learning: A Systematic Literature Review,” \emph{Applied Sciences}, vol. 12, no. 19, p. 9637, Sep. 2022, doi: 10.3390/app12199637.

\bibitem{ref94}
V. F. Rodrigues \emph{et al.}, “Fraud Detection and Prevention in E-Commerce: A Systematic Literature Review,” \emph{Electronic Commerce Research and Applications}, vol. 56, p. 101207, Oct. 2022, doi: 10.1016/j.elerap.2022.101207.

\bibitem{ref95}
R. Jáuregui-Velarde, L. Andrade-Arenas, P. Molina-Velarde, and C. Yactayo-Arias, “Financial Revolution: A Systemic Analysis of Artificial Intelligence and Machine Learning in the Banking Sector,” \emph{International Journal of Power Electronics and Drive Systems/International Journal of Electrical and Computer Engineering}, vol. 14, no. 1, p. 1079, Nov. 2023, doi: 10.11591/ijece.v14i1.pp1079-1090.

\bibitem{ref96}
W. Hilal, S. A. Gadsden, and J. Yawney, “Financial Fraud: A Review of Anomaly Detection Techniques and Recent Advances,” \emph{Expert Systems With Applications}, vol. 193, p. 116429, Dec. 2021, doi: 10.1016/j.eswa.2021.116429.

\end{thebibliography}

\end{document}